\documentclass{article}
\usepackage{microtype}
\usepackage{graphicx}
\usepackage{subcaption}
\usepackage{booktabs} % for professional tables

\usepackage{hyperref}

\usepackage{listings}
\usepackage[most]{tcolorbox}
\usepackage{listings}
\usepackage{enumitem}
\tcbuselibrary{listings, breakable, skins}

\usepackage[preprint]{icml2026}

\usepackage{amsmath}
\usepackage{amssymb}
\usepackage{mathtools}
\usepackage{amsthm}

\usepackage[capitalize,noabbrev]{cleveref}
\icmltitlerunning{Pitfalls in Evaluating Interpretability Agents}

\usepackage{amsmath,amsfonts,bm}

\def\eqref#1{equation~\ref{#1}}
\def\1{\bm{1}}

\DeclareMathAlphabet{\mathsfit}{\encodingdefault}{\sfdefault}{m}{sl}
\SetMathAlphabet{\mathsfit}{bold}{\encodingdefault}{\sfdefault}{bx}{n}

\definecolor{nicegreen}{rgb}{0.4, 0.615, 0.647}
\definecolor{niceyellow}{rgb}{0.95, 0.62, 0.3}
\definecolor{nicepurple}{rgb}{0.85, 0.5, 0.62}
\definecolor{myblue}{rgb}{0.298, 0.5, 0.9}
\definecolor{myred}{rgb}{1, 0.35, 0.35}
\definecolor{mygreen}{rgb}{0.2, 0.6, 0.2}
\definecolor{deeppink}{rgb}{1.0, 0.08, 0.58}

\usepackage{url}
\usepackage{graphicx}
\usepackage{graphicx}
\usepackage{booktabs}
\usepackage{array}

\begin{document}
\twocolumn[
  \icmltitle{
Pitfalls in Evaluating Interpretability Agents}

% All that glitters is not gold:\\ Pitfalls in the Evaluation of Interpretability Agents

% \small{\tamar{"Because We have Interpretability Agents, We Should Mind Their Limitations"
% \\
% "Because We Have Interpretability Agents, We Must Also Understand Their Limitations"
% \\
% "As Interpretability Agents Advance, So Do Their Limitations"
% \\
% "
% With Great Interpretability Agents Comes Great Responsibility
% "
% }}

% \small{\tamar{"Because We have Interpretability Agents, We Should and Need to be Evaluate Them"

%Automating Subnetworks Analysis with a Mechanistic Interpretability Agent

% Automating  Subnetworks Explanations with a Mechanistic Interpretability Agent

%Automating Mechanistic Explanations with an Interpretability Agent

%Automating Circuit Analysis with a Mechanistic Interpretability Agent

% Automating Algorithmic Descriptions of Language Models Operations with a Mechanistic Interpretability Agent

% Automating Algorithmic Descriptions of Language Models Operations with a Mechanistic Interpretability Agent

% Automating Causal explanations of Language Models

% Authors must not appear in the submitted version. They should be hidden
% as long as the \icmlfinalcopy macro remains commented out below.
% Non-anonymous submissions will be rejected without review.

\icmlsetsymbol{equal}{*}

\begin{icmlauthorlist}
    \icmlauthor{Tal Haklay}{Technion}
    \icmlauthor{Nikhil Prakash}{Northeastern}
    \icmlauthor{Sana Pandey}{MIT}
    \icmlauthor{Antonio Torralba}{MIT}
    \icmlauthor{Aaron Mueller}{Boston}
    \icmlauthor{Jacob Andreas}{MIT}
    \icmlauthor{Tamar Rott Shaham}{MIT}
    \icmlauthor{Yonatan Belinkov}{Technion,Kempner}
\end{icmlauthorlist}

  \icmlaffiliation{Technion}{Technion—IIT }
  \icmlaffiliation{Kempner}{Kempner Institute at Harvard University}
  \icmlaffiliation{Northeastern}{Northeastern University}
  \icmlaffiliation{Northeastern}{Northeastern University}
   \icmlaffiliation{Boston}{Boston University}
   \icmlaffiliation{MIT}{MIT}

  \icmlcorrespondingauthor{Tal Haklay}{tal.ha@campus.technion.ac.il}

  \icmlkeywords{Machine Learning, ICML}
\vskip 0.3in
]



\printAffiliationsAndNotice{}  % no special notice (required even if empty)

\begin{abstract}
Automated interpretability systems aim to reduce the need for human labor and scale analysis to increasingly large models and diverse tasks. Recent efforts toward this goal leverage large language models (LLMs) at increasing levels of autonomy, ranging from fixed one-shot workflows to fully autonomous interpretability agents. This shift creates a corresponding need to scale evaluation approaches to keep pace with both the volume and complexity of generated explanations. We investigate this challenge in the context of automated circuit analysis---explaining the roles of model components when performing specific tasks. To this end, we build an agentic system in which a research agent iteratively designs experiments and refines hypotheses. When evaluated against human expert explanations across six circuit analysis tasks in the literature, the system appears competitive. However, closer examination reveals several pitfalls of replication-based evaluation: human expert explanations can be subjective or incomplete, outcome-based comparisons obscure the research process, and LLM-based systems may reproduce published findings via memorization or informed guessing. To address some of these pitfalls, we propose an unsupervised intrinsic evaluation based on the functional interchangeability of model components. Our work demonstrates fundamental challenges in evaluating complex automated interpretability systems and reveals key limitations of replication-based evaluation.\footnote{Code available at \url{https://github.com/technion-cs-nlp/interp-agent}.}
\end{abstract}

\section{Introduction}

\begin{figure*}[t]
    \centering
    \includegraphics[width=0.95\linewidth]{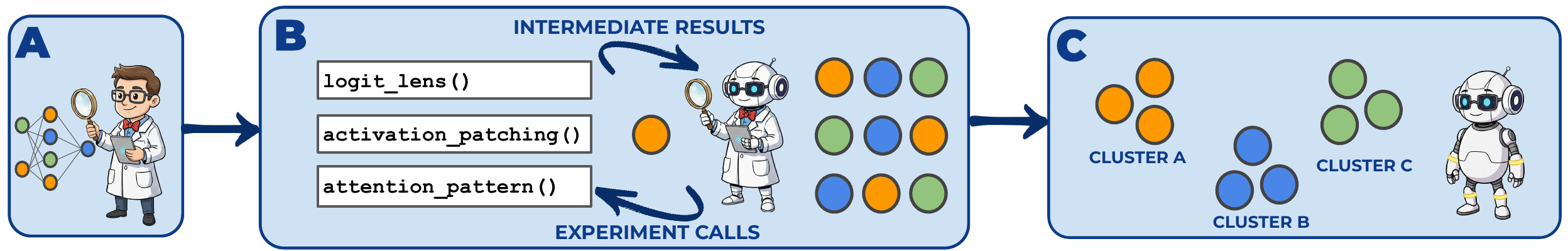}
    \vspace{0.4cm}

    \includegraphics[width=0.97\linewidth]{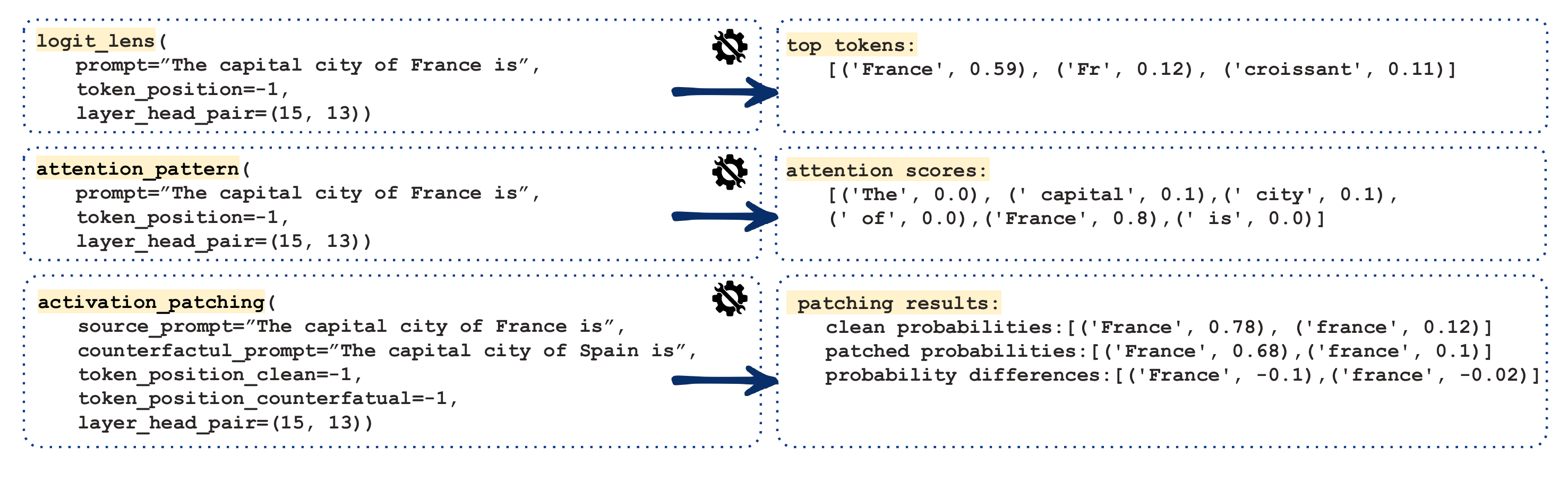}
\caption{\textbf{Top:} Overview of the system workflow.
(A) A researcher specifies a task and a circuit to analyze.
(B) A researcher agent iteratively analyzes each component independently, autonomously designing and running experiments as needed.
(C) Claude clusters components based on shared functionality inferred from the generated hypotheses. \textbf{Bottom:} Example of tool calls and the corresponding results returned by the tools. }

    \label{fig:agent_run}
    \vspace{-0.2cm}
\end{figure*}

Interpretability research aims to explain the internal computations of deep neural networks, such as how models implement in-context learning, solve arithmetic tasks, or track grammatical relationships across sentences \citep{olah-mech, wang2022interpretability, hanna2023does, todd2024function, davies2023discoveringvariablebindingcircuitry, prakash2024fine, nikankin2024arithmetic, todd2025incontextalgebra, feucht2025dualroute}. This work typically requires researchers to formulate hypotheses, design targeted experiments, and iteratively refine their conclusions—a process in which human expertise becomes a bottleneck. A growing body of work seeks to address this bottleneck by automating interpretability workflows \citep{hernandez2021natural, bills2023language, shaham2024multimodal, DBLP:conf/icml/PauloMJB25, DBLP:conf/acl/Gur-AriehMAGG25, minder2025narrowfinetuningleavesclearly}. Rather than relying on expert-crafted procedures, these systems execute analysis subroutines autonomously, often by leveraging AI models themselves. For example, LLMs are commonly used to interpret neurons or features by identifying patterns in the input tokens that activate them. In such approaches, human researchers define which experiments to run, while the LLM automates the analysis. Recently, \citet{shaham2024multimodal} demonstrated that autonomous interpretability agents can go further—designing experiments themselves and using the results to iteratively refine their hypotheses, much like human researchers. However, to trust such automated systems to produce accurate conclusions, they must be evaluated carefully. 

Evaluating the discoveries of autonomous interpretability systems is inherently challenging: their outputs are open-ended, can vary substantially across runs, and often lack clear ground-truth explanations. Moreover, equally valid explanations may differ in their level of abstraction or framing. Together, these factors make a scalable, automated evaluation particularly difficult. We investigate this evaluation challenge through the task of analyzing \emph{circuits} in language models---subnetworks of attention-heads and MLPs that implement specific computational pathways \citep{olah-mech, meng2022locating}. To understand how a model solves a task, researchers aim to identify these circuits, characterize the functional roles of individual components, and group components into clusters that jointly implement higher-level mechanisms \citep{wang2022interpretability, hanna2023does, prakash2024fine, merullo2023circuit, DBLP:conf/nips/Tigges0YB24, DBLP:conf/aistats/Garcia-Carrasco24}. While substantial prior work has focused on automating the \emph{discovery} of circuit components \citep{conmy2023towards, syed2310attribution, DBLP:conf/acl/HaklayOBMB25}, we focus on a less explored stage of the pipeline: assigning semantic, functional interpretations to the components of an already identified circuit for the task. This stage produces the human-interpretable explanations that describe how individual components contribute to the model’s behavior on the task. To this end, we study an \emph{agentic} system, in which a research agent operates iteratively—autonomously designing experiments and refining its analysis over multiple steps.

To assess whether such systems can replicate experts' research outcomes, we collect six prior circuit-analysis tasks from the literature and compare system outputs to explanations produced by human researchers \citep{meng2022locating, hanna2023does, DBLP:conf/nips/Tigges0YB24, prakash2024fine, merullo2023circuit}. Under this evaluation, our system appears competitive with prior analyses. However, closer examination reveals that apparent agreement with human-authored explanations masks deeper challenges in evaluating automated interpretability systems. Human-provided explanations may be incomplete or subjective; indeed, we identify cases in which our systems produce more accurate interpretations than those reported in prior work. Furthermore, Claude Opus~4.1 \citep{anthropic2025claude41}, which underlies our agent, exhibits evidence of memorization on at least one evaluated task and can infer others using superficial cues rather than experimental reasoning.

These failure modes point to broader limitations of replication-based evaluation. Evaluations that rely on human expert analyses are costly to obtain, can be subjective, and may reflect memorization rather than genuine understanding. To address these challenges, we propose an unsupervised intrinsic evaluation based on the functional interchangeability of model components, enabling assessment of component descriptions without reliance on expert judgment and at scale. While this metric does not address all the pitfalls discussed in this paper, it serves as an initial step toward scalable, unsupervised evaluation of automated circuit analysis systems.

To summarize, our work makes three main contributions: 
(i) a systematic analysis of an agentic system for circuit analysis; (ii) an exposition of key pitfalls in replication-based evaluation; and (iii) an unsupervised evaluation approach based on functional similarity.

\section{Automatic Circuit Analysis}
\label{sec:agent}

Our goal is to automate the procedure of circuit analysis: given a completion task (e.g., Indirect Object Identification (IOI);~\citealt{wang2022interpretability}), a set of task-specific prompts, and a list of model components participating in the task’s computation (i.e., the underlying circuit), we aim to produce a human-interpretable description of the functional roles of the circuit’s components. For example, such a description may explain what types of task-relevant information an attention head retrieves from the input, or what types of computations an MLP performs to facilitate correct prediction. Prior work has largely been performed by human researchers using interpretability techniques such as attention pattern analysis and causal interventions \citep{DBLP:conf/nips/VigGBQNSS20, wang2022interpretability}. This process is inherently iterative, typically requiring researchers to identify patterns across large volumes of experimental results and to formulate and test hypotheses using targeted prompts. Consequently, prior work in circuit analysis has typically required substantial time and the involvement of multiple researchers.

To automate this process, we develop an \emph{agentic} system centered on a research agent that autonomously and iteratively proposes hypotheses and designs experiments. Given a set of 20 task prompts \(P\) and a set of components \(C\) identified as belonging to the task’s circuit (e.g., attention heads or MLPs), the system outputs a collection of component clusters and their corresponding natural-language descriptions, \(\{(K_i, d_i)\}_{i=1}^{m}\), where each \(K_i \subseteq C\) groups components by shared functionality.

Figure~\ref{fig:agent_run} provides a high-level overview of the system, which follows a three-stage workflow: (i) a researcher specifies a task and a circuit to analyze; (ii) each circuit component is analyzed independently by a research agent to produce a task-specific functional description; and (iii) an LLM clusters the components based on shared functionality.

We next describe stages (ii) and (iii).\footnote{Given the progress in circuit discovery \citep{conmy2023towards, hanna2024have, DBLP:conf/icml/MuellerGWAABCFH25, ameisen2025circuittracing}, we assume stage (i) is given; that is, we assume access to the ground truth components.}

\subsection{Component-Level Analysis} \label{subsec:comp_analysis}
At this stage, the research agent produces a task-specific explanation for each of the circuit components independently. This analysis is grounded in experimental outputs from standard tools   that are widely used in  mechanistic interpretability work:
vocabulary projections, causal interventions, and attention patterns.  See Section \ref{subsec:tools}  for details.

 The \emph{research agent} is built on Claude Opus 4.1, guided by a system prompt (see App.~\ref{app:agent_system_prompt}) and equipped with the tools described below. Analogous to a human researcher, the agent operates iteratively. Each iteration of the agent follows a structured three-step process:  
(i) It begins by analyzing the experimental results obtained in the previous iteration.  
(ii) Based on this evidence, it articulates a set of $3$–$5$ plausible hypotheses that could explain the observed behavior.  
(iii) It then proposes the next set of experiments, specifying both the function calls and the rationale behind each experiment. For an example of a single agent iteration, see App.~\ref{ap:agent-iter}.

An experimental run concludes once the agent returns a \texttt{<final hypothesis>} tag without making any further function calls. The final output of the agent is a concise description of the component’s functionality, accompanied by a brief summary of key evidence supporting it (see App.~\ref{ap:final_hypothesis} for an example).

\subsection{Interpretability Tools}
\label{subsec:tools}
 Our tools are implemented with TransformerLens \citep{nanda2022transformerlens}, and described next.

% \begin{enumerate}
% \item \textbf
\noindent\textbf{Logit lens} \citep{nostalgebraist2020logitlens}: 
The \emph{logit lens} projects intermediate hidden states into the model’s output space using the final LayerNorm and the unembedding matrix. For a hidden state $h_t^{(l)}$ at layer $l$, the resulting  distribution over the vocabulary items is given by
$ \smash{p_t^{(l)} = \mathrm{softmax}(W_U\,\mathrm{LN} (h_t^{(l)}))}
$. 
This tool has been applied in numerous studies to analyze intermediate computations in language models \citep[][\it inter alia]{geva2021transformer, geva2022transformer, katz2024backward}.

\vspace{2pt}
% \item \textbf
\noindent\textbf{Activation patching} \citep{DBLP:conf/nips/VigGBQNSS20, geiger2020neural}: 
A causal intervention on a model component $c$ that measures how the model output would have changed if the output of $c$ were replaced with its value obtained from processing a different input. Given a clean prompt $\smash{x^{\mathrm{clean}}}$ and a counterfactual prompt $\smash{x^{\mathrm{cf}}}$, activation patching replaces the output of $c$ at a target position in the clean run with the corresponding activation from the counterfactual run. Such tools have been instrumental in numerous interpretability studies \citep{wang2022interpretability, meng2022locating, prakash2024fine}.

% \vspace{-2pt}
% \item \textbf{
\paragraph{Attention patterns:}
Given a prompt, a token position, and an attention head, this function returns the corresponding attention pattern. Such attention measures are simple yet useful tools that help researchers form hypotheses about the roles of attention heads \citep{clark2019does, DBLP:conf/nips/VigGBQNSS20, wang2022interpretability}.   To mitigate attention sink effects, we ignore the BOS token position and re-normalize the attention weights \citep{vig2019analyzing}.

\vspace{-2pt}
% \item \textbf
\paragraph{Token Positions:}  
Given a prompt, this function returns the tokenized prompt along with token indices.  This tool allows the agent to identify which token positions to target when running experiments, and is particularly useful when the agent generates new prompts for analysis.

The interaction with these tools is mediated through a structured interface: at each iteration, the agent specifies which tools to invoke, and their corresponding arguments, including generating input prompts (e.g., designing prompt pairs for activation patching). Then, an external system executes the requested functions, and the outputs are returned to the agent for subsequent analysis. 
As an initialization, we precompute attention-pattern and logit-lens of 10 task prompts and include the outputs in the prompt to the agent.

\subsection{Clustering Components by Shared Functionality}

Prior work has shown that multiple components within a circuit may implement similar functionality and can therefore be grouped into clusters, enabling higher-level summaries of model behavior \citep{wang2022interpretability, hanna2023does, prakash2024fine}. Moreover, such clusters typically consist of components that operate at the same token position. After the agent produces a final description for each component, we perform a second analysis in which these component-level descriptions are aggregated and clustered by functional similarity. This stage does not involve any additional experiments. Instead, we prompt Claude Opus~4.1 to cluster components based on their final functional role descriptions and the associated evidence (see App~\ref{ap:cluatering}). Following prior work, clustering is performed separately for each token position. The outcome of this stage is a set of clusters, each accompanied by a description of the shared functionality of components in it.

\begin{figure}[t]
    \centering
\includegraphics[width=\linewidth]{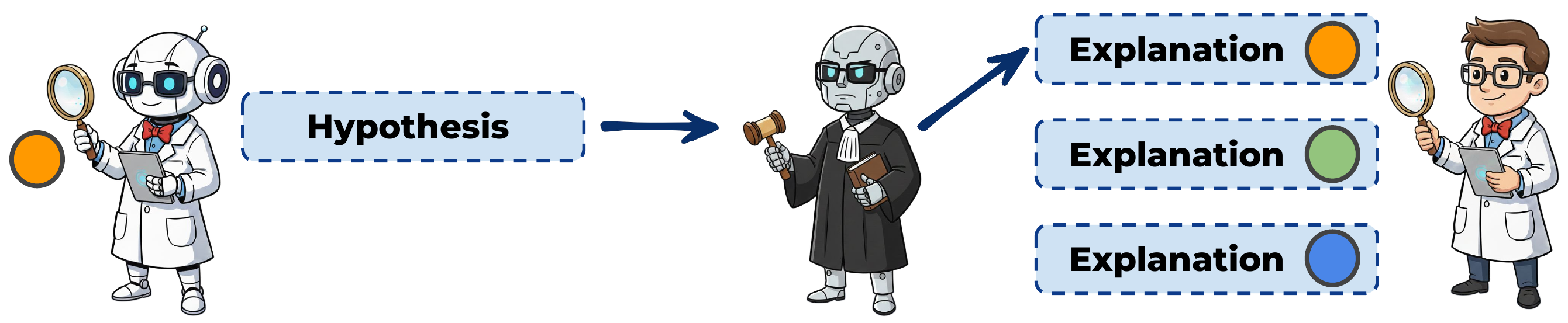}

    \caption{The judge workflow. The system produces hypotheses explaining the functionality of individual components and clusters. A judge model is tasked with matching the hypothesis to one of the descriptions reported by the researchers in the original paper.}
    \label{fig:judge}
\end{figure}

\section{Extrinsic Evaluation with Human Explanations}
\label{sec:extrinsic_evaluation}

Evaluating explanations is a core and long-standing challenge in interpretability research \citep{doshi2017towards,jacovi2020towards,schwettmann2023find,huang2023rigorously}. 
While human judgments are often treated as the gold standard, manual evaluation introduces a substantial scalability bottleneck, making it impractical for automated interpretability research. We therefore seek an evaluation protocol that is both scalable and automated, allowing explanations to be assessed without human supervision.

A common approach for evaluating research agents is to test whether they replicate findings from published work. This paradigm underlies benchmarks across domains including machine learning engineering \citep{chan2024mle}, astrophysics \citep{DBLP:journals/corr/abs-2510-24591}, and scientific discovery more broadly \citep{DBLP:conf/icml/StaraceJSACMDMK25}. In our setup, we test whether our system can reproduce explanations produced by human researchers. We identify six  studies that performed circuit analysis, %provide all the information necessary for such a comparison, 
including explicit descriptions of components' functionalities and corresponding clusters \citep{wang2022interpretability, hanna2023does, prakash2024fine, merullo2023circuit, DBLP:conf/nips/Tigges0YB24, DBLP:conf/aistats/Garcia-Carrasco24}. Table~\ref{tab:circuits-summary} provides a summary of the task setups. Additional details for each task are provided in App.~\ref{ap:tasks}.

\begin{table}[h]
\caption{Summary of tasks from prior circuit analyses. Each cluster represents a group of components with shared functionality.}
\centering
\footnotesize
\resizebox{\columnwidth}{!}{%
\begin{tabular}{llcc}
\toprule
\textbf{Task} & \textbf{Model} & \textbf{Circuit Size} & \textbf{\#Clusters} \\
\midrule
\textbf{IOI} & 
GPT-2-Small  & 
18 Heads & 6  \\
\addlinespace
\textbf{IOI} & 
Pythia-160M  & 
14 Heads  & 7 
 \\
\addlinespace
\textbf{Greater-Than } & 
GPT-2-Small  & 
8 Heads, 4 MLPs & 2 \\
\addlinespace
\textbf{Acronyms}  & GPT-2-Small
 & 8 Heads & 3 
\\
\addlinespace
\textbf{Colored Objects } & 
GPT-2-Medium  & 
27 Heads  & 3   \\
\addlinespace
\textbf{Entity Tracking} & 
LLaMA-7B  & 
64 Heads & 4 \\
\bottomrule
\end{tabular}%
}
\label{tab:circuits-summary}
% \vspace{-5pt}
\end{table}

\begin{figure*}[t]
    \centering
    % Top row
    \includegraphics[width=0.95\linewidth]{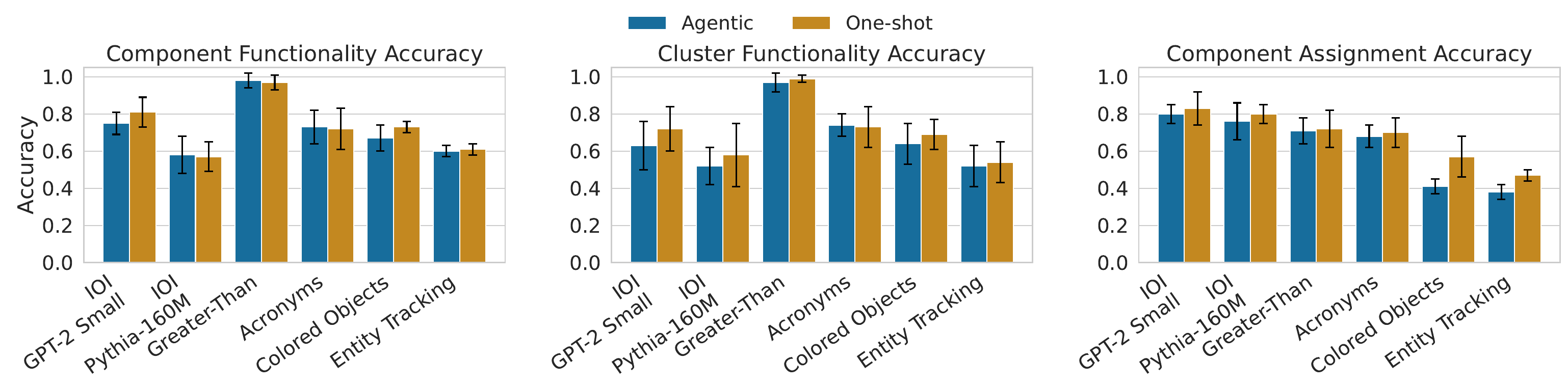}
    \caption{Performance comparison  across six circuit analysis tasks. Left: Component Functionality Accuracy measures how well individual component explanations match human-labeled clusters. Middle: Cluster Functionality Accuracy measures how well cluster explanations match human-labeled clusters. Right: Component Assignment Accuracy assesses cluster alignment with expert-defined clusters using optimal matching. While the systems obtain relatively high results,  they usually do not perfectly match  expert explanations.  }
    \label{fig:comparison_methods}
\end{figure*}

 \subsection{Metrics}
To measure how well agent-generated explanations match those reported in prior work, we use an LLM-as-a-judge approach. A judge model (GPT-5) receives an explanation from the agent and is tasked with matching it with one human explanation out of all possible human explanations for the circuit clusters (see Figure~\ref{fig:judge}). The judge is allowed to assign multiple agent-generated explanations to the same human-generated explanation. We repeat this process for two types of the agent's explanations; first, we evaluate explanations of each component (stage ii), then we evaluate the agent's explanations for the functionality of each cluster (stage iii).

We evaluate performance using three metrics:

 \textbf{Component Functionality Accuracy.}
The proportion of individual component-level explanations that the judge model correctly matches to their human-labeled clusters.

\textbf{Cluster Functionality Accuracy.} 
The proportion of components in each agent-generated cluster that the judge assigns to the correct human-labeled cluster.
Formally, let $K = \{K_1, \dots, K_m\}$ denote the set of clusters produced by the agent. We compute the fraction of the correctly assigned components in each cluster $K_j$ and average across clusters:
\vspace{-5pt}
\begin{equation}
\begin{aligned}
\text{Accuracy}
=
\frac{1}{m}
\sum_{j=1}^{m}
\frac{1}{|K_j|}
\sum_{c \in K_j}
\mathbf{1}\!\left[\hat{y}(c) = y(c)\right]
\end{aligned}
\end{equation}

\textbf{Component Assignment Accuracy.} This metric measures how well the predicted clusters align with expert-defined clusters from prior work and does not involve the judge. Since cluster labels are arbitrary (the system might call a cluster ``A'' while the paper calls it ``Name Movers''), we cannot directly compare assignments. Instead, we find the optimal one-to-one matching between predicted and expert clusters with the Hungarian algorithm \citep{kuhn1955hungarian}, which maximizes the number of components that end up in corresponding clusters. We then report the fraction of components correctly assigned under the optimal matching.

We report both Component Assignment Accuracy and Cluster Functionality Accuracy, as they evaluate different aspects of agreement: the former assesses agreement on which components share a function, while the latter assesses agreement on what that function is.

\subsection{Does Greater Autonomy Improve Performance?}

As a baseline to our agentic system, we also examined a common paradigm in automated interpretability. Rather than using a \emph{research agent} to analyze each component, we provide an LLM with experimental outputs as static input and ask it to produce interpretations in a single pass. This approach follows prior work in which LLMs are used to interpret  activation patterns of neurons or sparse autoencoder (SAE) features  \citep{bills2023language, DBLP:conf/icml/PauloMJB25}.

In this setup, which we refer to as the \emph{one-shot} system,  we select 10 of the 20 prompts provided in the system prompt and apply all available interpretability tools described in Sec.~\ref{subsec:tools}. For patching experiments, we adopt a simple strategy by randomly sampling 10 counterfactual prompts from the task dataset. All resulting experimental outputs are then provided to Claude Opus~4.1 via the system prompt, which analyzes them and produces a description of each component’s functional role.

\subsection{Results}
For each task, we run the component-level analysis three times and perform clustering five times per analysis, yielding fifteen clusterings; all reported results are averaged across runs. In total, the agent issued 6{,}382 tool calls, averaging $14.2$ calls and $4.5$ iterations per component analysis. 

Figure~\ref{fig:comparison_methods} presents the evaluation results. While neither system consistently matches human-level analyses, both the agentic and one-shot systems achieve competitive performance. Notably, \emph{increased autonomy does not necessarily lead to consistent performance gains}: the agentic system performs comparably to the one-shot baseline. We further analyze the comparable performance in Section~\ref{subsec:outcome_eval}.

\begin{figure*}[t]
    \centering

    \includegraphics[width=0.9\linewidth]{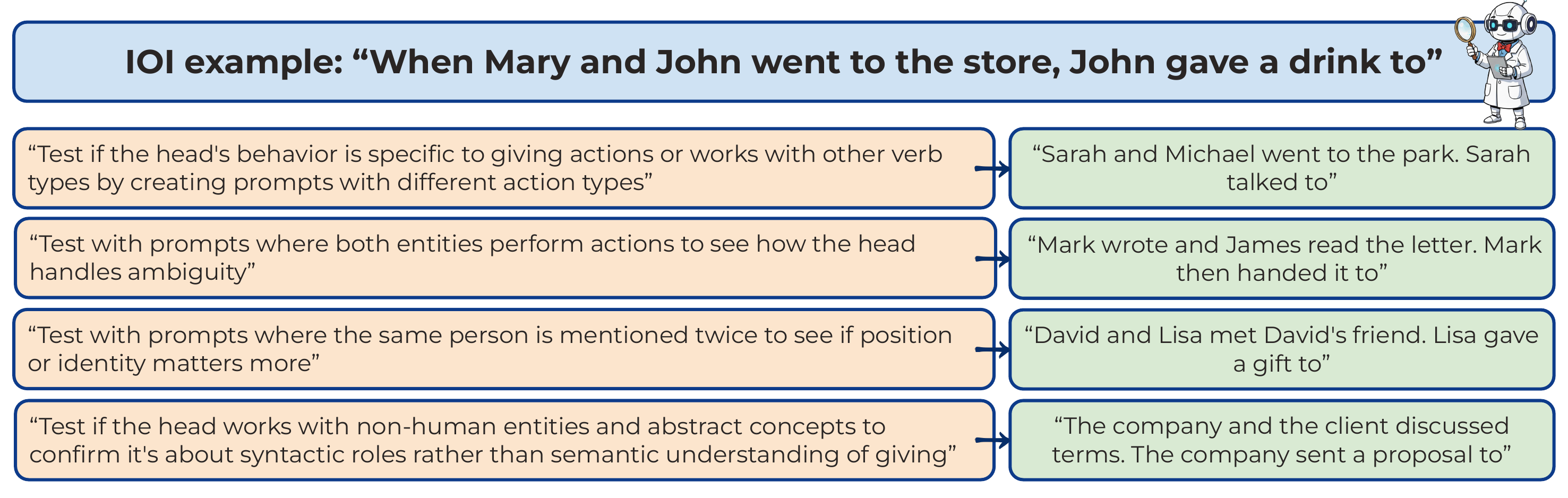}

    \caption{Examples of the agent’s experimental designs.
\textcolor{blue}{Blue}: an IOI task example from the initial prompt set provided to the agent.
\textcolor{orange!80!black}{Orange}: the agent’s motivation for the experiment.
\textcolor{green!60!black}{Green}: new example prompts proposed by the agent to test its hypotheses.}
    \label{fig:agent_hypothesis}
\end{figure*}
\section{Pitfalls in Evaluating Automated Interpretability Systems}
\label{sec:Challenges}

While the evaluation framework in Section~\ref{sec:extrinsic_evaluation} is intended to assess whether the two systems can reproduce human circuit analyses, closer examination reveals several important limitations. We examine these limitations in detail below.

\subsection{Findings in Prior Work May Be Ambiguous}

Disagreements between the system’s hypotheses and published descriptions do not always reflect model error. In several cases, they arise from subjectivity or incompleteness in prior human interpretations. For example, attention head 2.6 in the IOI task on Pythia-160M  had been described as a previous-token head \citep{DBLP:conf/nips/Tigges0YB24}, yet the agent’s experiments did not consistently reveal a previous-token attention pattern. To test this, we conducted a broader evaluation over 150 examples from the PILE dataset \citep{gao2020pile}, The previous token received the highest attention score in only 42\% of cases, raising the question of whether this head should really be labeled a “previous-token” head. Similarly, in the Entity Tracking task \citep{prakash2024fine}, which uses prompts of the form “The apple is in box C, the cat is in box D, … Box C contains the”, a large group of components is labeled as value fetcher heads, responsible for retrieving the value associated with the correct object. In contrast, both systems identify several heads whose behavior does not consistently match the description in the original paper. For further analysis, see App.~\ref{ap:in-comp}. These cases highlight a broader challenge in circuit analysis: observed behaviors are not always consistent across inputs, and when a component exhibits a pattern in only a subset of test cases, assigning it a single functional explanation can become ambiguous.

\subsection{Outcome-Based Evaluation ignores Research Process}
\label{subsec:outcome_eval}
Our evaluation compares final explanations against those reported in prior work. However, this outcome-based comparison discards information about the research process itself: two researchers may arrive at the same conclusion through fundamentally different investigative paths. Figure~\ref{fig:agent_hypothesis} shows that the agent engages in substantive hypothesis generation and experimental design, with particular emphasis on testing the generalization of its hypotheses and refining them by evaluating prompts that differ from those provided initially. In contrast, the one-shot baseline is unable to refine its hypothesis or explore alternative explanations. We view this as a key advantage of the agentic system over the one-shot baseline. Yet the current evaluation framework provides no direct mechanism to reward this behavior.

\subsection{Are We Actually Measuring Reasoning?}
\subsubsection{Signs Of Memorization}

\begin{figure}[t]
    \centering
    \includegraphics[width=0.48\textwidth]{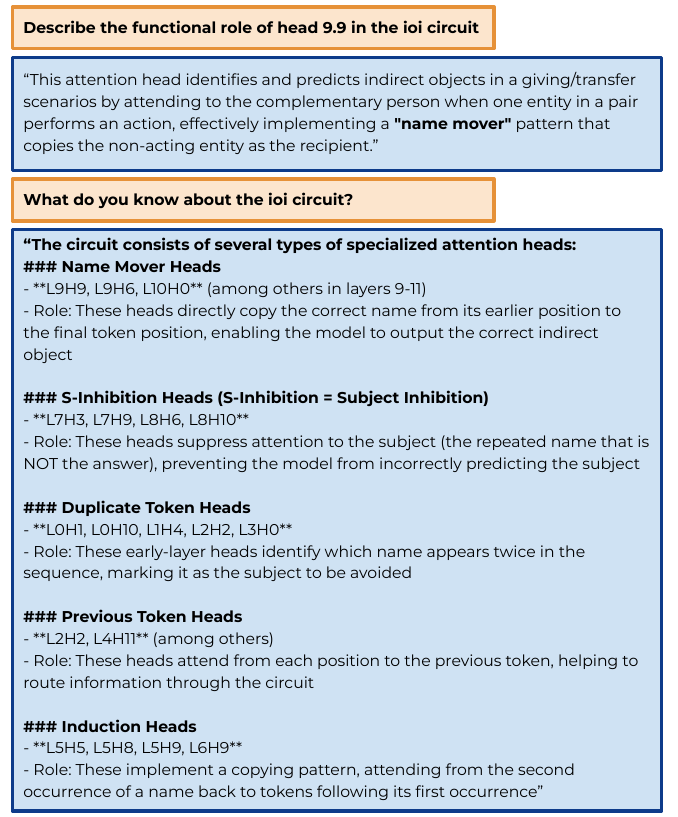}
\caption{\textbf{Top:} Example of indirect memorization in a final hypothesis produced by the one-shot system for attention head 9.9 in the IOI task. The system explicitly uses the term ``name mover head'', reproducing the terminology introduced in the original IOI paper. For more details on ``name mover'' heads, see App.~\ref{ap:task-ioi}.\\
\textbf{Bottom:} Example of direct recall of the IOI circuit by Claude when it is explicitly asked to recall the circuit from memory. For more details on the prompt used and the full response, see App.~\ref{ap:ioi_memory}.}
\label{fig:memorization}
\end{figure}

A key concern in using LLMs such as Claude Opus 4.1 as the backbone for both systems is the possibility that the model memorized the published findings during training\footnote{We use Claude through API without retrieval capabilities, so there is no risk of it recalling information from retrieved sources. Systems with search capability would need to account for that too.} \citep{DBLP:conf/uss/CarliniTWJHLRBS21, DBLP:conf/iclr/CarliniIJLTZ23}. By directly prompting Claude to list all circuit's components and their functionalities (without running any experiments), we confirmed that it has memorized at least the IOI circuit, including the exact components (e.g., layer and head indices) and the functionality attributed to each head (See App.~\ref{ap:ioi_memory}). Additionally, GPT-5, the model used as our judge, also showed signs of direct recall of the IOI task. See Figure~\ref{fig:memorization} and App.~\ref{ap:ioi_memory} for concrete examples.

Our experience suggests that, without explicit tests for memorization, such phenomena are easy to overlook:
    (i) During analysis, Claude did not state that it recognized the IOI task; (ii) Most generated explanations did not reuse the terminology employed in the original paper; and (iii) most surprisingly, both systems' reproductions were \emph{not} perfect, despite effectively knowing the researchers’ findings.

Even in other tasks where Claude did not explicitly acknowledge familiarity with published findings, it was often able to infer plausible functionalities that closely resembled those reported in the papers. For instance, it was able to “guess” a high-level algorithm similar to that identified in the Entity Tracking circuit (see App.~\ref{ap:et_memory}). We acknowledge that this issue may be amplified in our experimental setting, where only a small set of well-established behaviors are documented in the literature. Consequently, the hypothesis space is relatively small, making it easier for the agent to arrive at accurate guesses without conducting any experiments.

We argue that these observations have important implications for future research on automated interpretability. \textbf{Comparisons between research agents and published work should be made with care, explicitly accounting for memorization and informed guessing.} The limited number of well-studied circuit analysis tasks further amplifies these risks, making replication-based evaluations particularly vulnerable to memorization and overfitting.

\subsubsection{Disentangling Memorization from Reasoning}

We further investigate whether Claude’s performance in the IOI task reflects genuine reasoning grounded in experimental evidence, or whether it reflects memorization.

To disentangle memorization from reasoning, we progressively add noise to the experimental results and examine the resulting impact on system performance. If Claude does not rely on the experimental evidence, we would expect only limited degradation as noise increases. For each tool (Section \ref{subsec:tools}), we generate a “noised” version of the original results by randomly permuting them. For example, in the logit-lens experiments, we randomize the probabilities assigned to each token. See App.~\ref{ap:noise} for further details. We then construct a noisy result using a mixing parameter $\alpha \in [0,1]$:
\vspace{-6pt}
\begin{equation}
\begin{aligned}
(1 - \alpha)\cdot \text{original\_result} + \alpha \cdot \text{permuted\_result}.
\end{aligned}
\end{equation}

We run both systems on the IOI task for $\alpha \in [{0, 0.5, 0.6, 0.7, 0.8, 0.9, 1}]$, repeating each experiment three times with different seeds. 

\begin{figure}[t]
    \centering
    \includegraphics[width=0.45\textwidth]{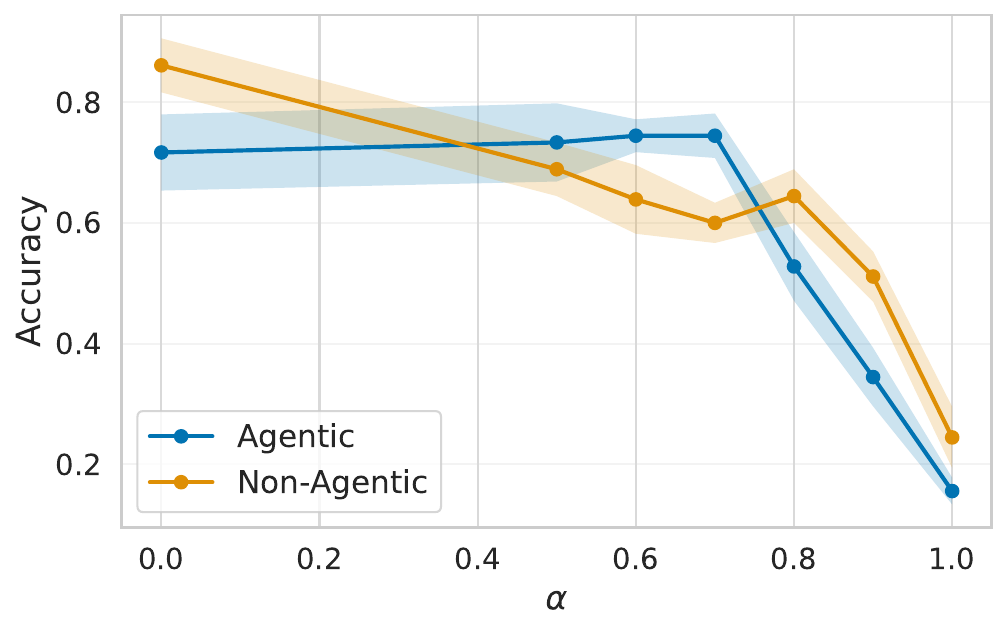}
    \caption{Each point represents the mean component functionality accuracy, averaged across all components in the IOI task, at a given noise level $\alpha$. Both systems exhibit clear performance degradation as noise increases. Running the experiments with additional random seeds yields similar trends (see App.~\ref{ap:noise}).}
    \label{fig:noise_exp}
\end{figure}

Results are shown in Figure~\ref{fig:noise_exp}.  When $\alpha > 0.7$, both systems exhibit substantial performance degradation, suggesting that their behavior relies more heavily on experimental evidence than on memorization. 
At lower $\alpha$ values, we observe  minimal performance degradation. This may either reflect memorization or suggest the system is able to recover the underlying structure of the original results despite the noise.
Overall, these results suggest that memorization alone cannot fully explain system behavior across noise regimes.

\section{Intrinsic Evaluation Without Ground Truth Explanations}
In this section, we present an alternative, fully automated evaluation approach that does not rely on expert explanations.

The core idea is to assess how \emph{coherent} (or \emph{tight}) the discovered clusters are: if a system groups together components that truly share an underlying function, then components within the same cluster should exhibit similar behavior, whereas components in different clusters should behave differently. To evaluate this, we require a distance function that places functionally similar components close together and functionally distinct components far apart. Rather than relying on subjective human judgments, we define a model-grounded, behavior-based measure of component similarity via \emph{swap-invariance}: if two components implement the same underlying function, then swapping one component's weights with those of the other should leave the model's behavior largely unchanged.

\paragraph{Implementation details.}
Concretely, we focus on attention heads.
For a given pair of heads, we consider swaps of their KQ circuits and separately their OV circuits \citep{olah-mech}. The OV circuit, defined by the value and output projection matrices $(W_V, W_O)$, determines \emph{what} information the head writes to the residual stream. The KQ circuit, defined by the key and query projection matrices $(W_K, W_Q)$, determines \emph{which token positions} the head attends to. For each pair of attention heads, we separately swap their KQ and OV circuits and measure the resulting change in the model’s output behavior.

\begin{table*}[t]
\caption{Intrinsic cluster quality. Silhouette-based cluster quality scores for random, expert-defined, agentic, and one-shot clusterings across tasks. For agentic and one-shot systems, we report the best and mean scores across runs. The final column reports the Kendall rank correlation ($\tau$) and associated $p$-value between intrinsic cluster quality and component assignment accuracy.}
\centering
\small
% \scriptsize
% \setlength{\tabcolsep}{8pt}
% \resizebox{\columnwidth}{!}{%
\begin{tabular}{l c c cc cc r}
\toprule

& \textbf{Random}
& \textbf{Expert}
& \multicolumn{2}{c}{\textbf{Agentic}}
& \multicolumn{2}{c}{\textbf{One-shot}}
& \textbf{Kendall ($\tau,p$)} \\
\cmidrule(lr){4-5} \cmidrule(lr){6-7}
\textbf{Task} & &
& \textbf{Mean} & \textbf{Best}
& \textbf{Mean} & \textbf{Best}
& \\
\midrule
IOI
& -0.28 $\pm$ 0.07
& 0.21
& 0.03 $\pm$ 0.09
& 0.12
& -0.08 $\pm$ 0.13
& 0.21
& (0.739, 0.00) \\
IOI (Pythia-160M)
& -0.24 $\pm$ 0.08
& 0.10
& 0.07 $\pm$ 0.10
& 0.15
& 0.00 $\pm$ 0.08
& 0.09
& (0.034, 0.81) \\
% Greater-Than
% & -0.03 $\pm$ 0.13
% & ---
% & 0.36
% & 0.08 $\pm$ 0.15
% & 0.36
% & 0.22 $\pm$ 0.17
% & --- \\
Acronyms
& -0.18 $\pm$ 0.13
& 0.16
& -0.02 $\pm$ 0.17
& 0.30
& 0.00 $\pm$ 0.10
& 0.30
& (0.506, 0.02) \\
Colored-Objects
& -0.28 $\pm$ 0.07
& 0.02
& 0.01 $\pm$ 0.07
& 0.05
& 0.02 $\pm$ 0.05
& 0.06
& (0.318, 0.02) \\
Entity-Tracking
& -0.24 $\pm$ 0.01
& 0.17
& -0.13 $\pm$ 0.02
& -0.09
& -0.09 $\pm$ 0.02
& -0.08
& (0.378, 0.00) \\
\bottomrule
\end{tabular}
% }
\label{tab:circuit-recovery-scores}
\end{table*}

\paragraph{Evaluation metrics.}
To quantify the effect of each swap, we measure the Jensen--Shannon distance (i.e., the square root of the Jensen--Shannon divergence) between the model’s original next-token prediction distribution and the distribution obtained after the swap. We then define the distance between two attention heads $h_1$ and $h_2$ as the average of the distances induced by swapping their KQ and OV circuits:
\vspace{-4pt}
\begin{equation}
\begin{aligned}
\mathrm{dist}(h_1,h_2) &= \frac{1}{2}\Big(
\sqrt{\mathrm{JSD}_{\mathrm{KQ}}(h_1,h_2)} \\
&\qquad + \sqrt{\mathrm{JSD}_{\mathrm{OV}}(h_1,h_2)}
\Big)
\end{aligned}
\end{equation}

 We measure the distance between every pair of heads within each task by running these swap experiments on 40 task examples. This distance yields a fully unsupervised, model-grounded measure of functional similarity between attention heads, independent of any expert explanations. Using this distance, we compute the average silhouette score under a given clustering assignment, which serves as an intrinsic measure of cluster quality.

We evaluate cluster quality across four clustering types: (1) clusters produced by the agentic system, (2) clusters produced by the one-shot system, (3) clusters reported in the original papers (referred to as expert clusters), and (4) random clusterings. For the latter, we randomly assign circuit components to clusters while matching the cluster-size distributions of the system-generated clusterings.

\paragraph{Results.}
Table~\ref{tab:circuit-recovery-scores} reports the results of this analysis. As expected, random clusterings yield low, negative silhouette scores, serving as a baseline for this analysis. 
 Expert clusters achieve much higher scores. Both the agentic and one-shot systems achieve silhouette scores close to zero on average, but still above the random baseline. However,  some individual clustering assignments attain relatively high scores, suggesting that while average performance under this evaluation is modest, the systems occasionally discover meaningful structure in the circuit representations.

\begin{figure}[h]
    \centering

    \includegraphics[width=0.97\linewidth]{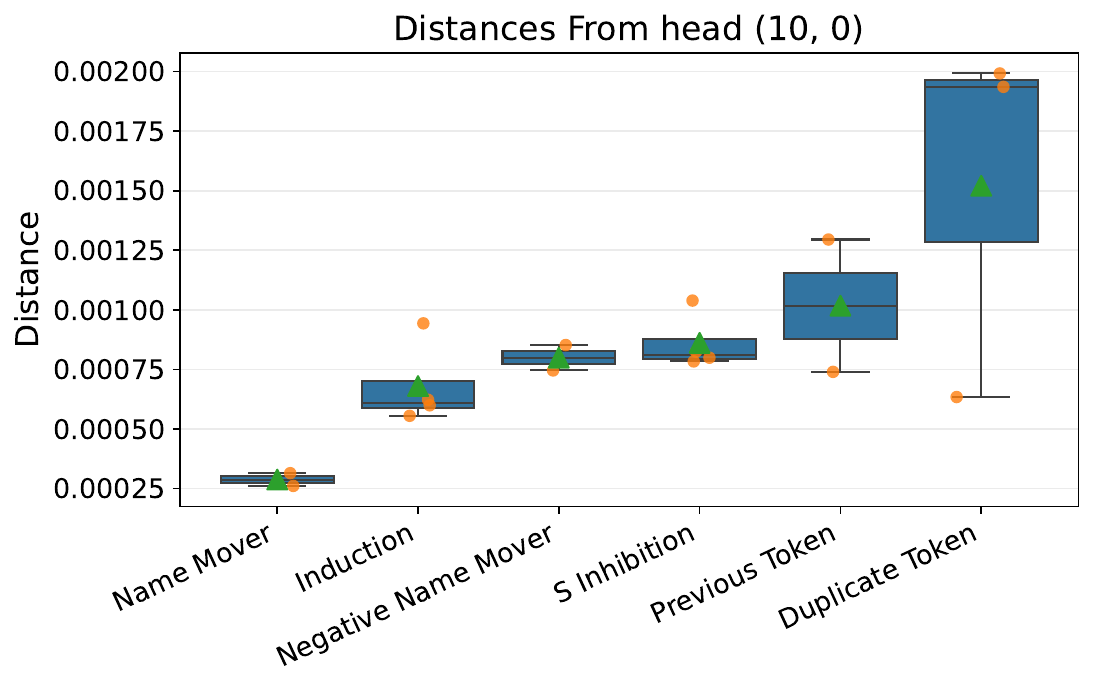}

    \caption{Swap-based distances from attention head $(10,0)$ (a \textsc{Name Mover} head) to all other heads in the IOI circuit, grouped by the reported functional clusters. Other \textsc{Name Mover} heads are closer to head $(10,0)$ than heads from different clusters.}

    \label{fig:10_0_distence}
\end{figure}

We additionally compute the Kendall rank correlation between our intrinsic metric and Component Assignment Accuracy, which measures agreement with clusters reported in prior work. For all tasks except the IOI Pythia-160, we observe positive correlations, indicating that clusters that are more coherent under our intrinsic, unsupervised metric often align with expert-defined clusters. While the strength of this relationship varies across tasks, these results suggest that the proposed intrinsic measure captures meaningful structure consistent with prior analyses. Figure~\ref{fig:10_0_distence} illustrates this consistency. We plot the swap-based distances from attention head $(10,0)$, identified in prior work as a \textsc{Name Mover} head, to all other heads in the IOI circuit. Heads within the \textsc{Name Mover} cluster exhibit systematically smaller swap-induced distances than heads from other clusters.

Notably, both the extrinsic and the intrinsic evaluations exhibit substantial run-to-run variance due to the stochasticity of the underlying model. The proposed intrinsic metric provides an unsupervised measure of cluster quality, allowing practitioners to compare multiple runs and select more coherent clustering without relying on ground-truth explanations.

We emphasize that this approach does not address all of the pitfalls discussed in Section~\ref{sec:Challenges}. Accordingly, it should be viewed as a proof of concept for unsupervised evaluation.

\section{Conclusions}

As automated interpretability systems become more autonomous and widely deployed, evaluation methodologies must evolve to keep pace with their capabilities. In this work, we show that commonly used replication-based evaluations—comparing agent-generated explanations to prior human analyses—can be systematically misleading. Through empirical analysis, we identify several failure modes of this paradigm, including reliance on incomplete or subjective ground-truth explanations, insensitivity to the reasoning process by which explanations are produced, and vulnerability to memorization or informed guessing by LLMs.

To begin addressing these challenges, we propose an unsupervised intrinsic evaluation framework based on functional interchangeability of model components. By grounding evaluation directly in model behavior, our approach enables scalable assessment of cluster coherence without relying on human evaluation or prior expert explanations. While this metric is limited in scope and does not resolve all identified pitfalls, our results suggest that it captures meaningful functional structure and complements replication-based evaluations.

Taken together, our findings highlight the central role of evaluation in the design of automated interpretability systems. As such systems take on increasingly open-ended scientific roles, effective evaluation will require methodologies that are robust to memorization, sensitive to reasoning processes, and grounded in the behavior of the underlying models. We view this work as a step toward such evaluation frameworks and as a call for more principled benchmarks for agentic interpretability.
% \newpage
\section*{Impact Statement}

This work aims to advance the evaluation of automated interpretability systems, which are increasingly used to analyze and explain the behavior of machine learning models. Reliable interpretability is important for transparency and oversight, particularly as models are deployed in high-stakes settings.

Our findings highlight that commonly used replication-based evaluation frameworks can obscure important failure modes, including memorization of prior analyses, reliance on superficial cues, and sensitivity to subjective or incomplete ground-truth explanations. These issues could lead to overconfident or misleading conclusions about the capabilities and trustworthiness of automated interpretability systems.

By identifying these pitfalls, our work aims to support the development of more robust and reliable evaluation methodologies. We hope these insights will help future research anticipate evaluation failures, design stronger benchmarks, and apply automated interpretability tools more cautiously and responsibly.

\section*{Acknowledgments}

This research was supported by the Israel Science Foundation (grant No.\ 2942/25), and the European Union (ERC, Control-LM, 101165402). Views and opinions expressed are those of the authors only and do not necessarily reflect those of the European Union or the European Research Council Executive Agency; neither the European Union nor the granting authority can be held responsible for them.
Additional support was provided by the ARL grant \#W911NF-24-2-0069, the MIT-IBM Watson AI Lab grant \#W1771646, Hyundai Motor Company, and the DARPA AIQ program through the DARPA CMO contract number HR00112520025.
TH, NP, AM, and YB were supported by Coefficient Giving. TH is also supported by the Council for Higher Education (VATAT) Scholarship for Outstanding Women Doctoral Students in Hi-Tech. SP is supported by the National Science Foundation Graduate Research Fellowship Program under Grant No. (2141064). Any opinions, findings, and conclusions or recommendations expressed in this material are those of the author(s) and do not necessarily reflect the views of the National Science Foundation.

% \newpage

\bibliography{icml2026_conference}
\bibliographystyle{icml2026}

\newpage
\appendix
\appendix
\onecolumn

\textbf{{\LARGE{Appendix}}}

% ============ LLM PROMPT BOXES ============

% System prompt box
\newtcolorbox{systemprompt}[1][]{
  colback=gray!5,
  colframe=gray!70,
  fonttitle=\bfseries,
  title=System Prompt,
  breakable,
  #1
}

% User message box
\newtcolorbox{userprompt}[1][]{
  colback=blue!5,
  colframe=blue!50,
  fonttitle=\bfseries,
  title=User,
  breakable,
  #1
}

% Assistant response box
\newtcolorbox{assistantprompt}[1][]{
  colback=green!5,
  colframe=green!50,
  fonttitle=\bfseries,
  title=Assistant,
  breakable,
  #1
}

% ============ CODE BOXES ============

% Generic code style
\lstdefinestyle{mystyle}{
  basicstyle=\ttfamily\small,
  breaklines=true,
  breakatwhitespace=false,
  numbers=left,
  numberstyle=\tiny\color{gray},
  numbersep=8pt,
  showstringspaces=false,
  tabsize=2,
  keepspaces=true,
}

% Python code box
\newtcblisting{pythoncode}[1][]{
  listing only,
  listing style=mystyle,
  colback=yellow!5,
  colframe=orange!70,
  fonttitle=\bfseries,
  title=Python,
  breakable,
  left=6mm,
  #1
}

% Generic code box (specify language in options)
\newtcblisting{codebox}[2][]{
  listing only,
  listing style=mystyle,
  listing options={language=#2, style=mystyle},
  colback=gray!5,
  colframe=gray!70,
  fonttitle=\bfseries,
  title=#2,
  breakable,
  left=6mm,
  #1
}

\section{Tools}
Instructions for invoking and interpreting the interpretability tools are provided to both systems through their respective system prompts.

\begin{pythoncode}[title=ModelInvest Class, breakable]
class ModelInvest:
    
    def __init__(self, model_name: str, hf_name: str = None, hf_tokenizer: str = None):
        """
        Initialize the ModelInvest class with a given model.
        
        Args:
            model_name: Name of the model in TransformerLens
            hf_name: Optional HuggingFace model name for loading via HF
            
        Example:
            >>> # Initialize with TransformerLens model only
            >>> model_invest = ModelInvest("gpt2-small")
            >>> 
            >>> # Initialize with both TransformerLens and HuggingFace model
            >>> model_invest = ModelInvest("llama-7b", "meta-llama/Llama-2-7b")
        """
        model, tokenizer = None, None
        if hf_name and hf_tokenizer:
            model = AutoModelForCausalLM.from_pretrained(
                hf_name,
                torch_dtype=torch.float32,  
                device_map="cpu"            
            )
            tokenizer = AutoTokenizer.from_pretrained(hf_tokenizer)
            tokenizer.pad_token_id = tokenizer.eos_token_id
            tokenizer.padding_side = "right"
            tokenizer.add_bos_token = False
            
        model = HookedTransformer.from_pretrained(
            model_name=model_name,
            hf_model=model,
            tokenizer=tokenizer,
            center_writing_weights=False,
            center_unembed=False,
            trust_remote_code=True,      
            fold_ln=False,
            device_map="auto")
        model.set_use_attn_result(True)
        #model.set_use_attn_in(True)
        #model.set_use_hook_mlp_in(True)
        #model.set_use_split_qkv_input(True)
        self.model = model
        self.model_name = model_name
        self.prepend_bos=True


    def get_token_indices_in_prompt(self, prompts: List[str]) -> List[Tuple[str, int]]:
        """
        Get tokens and their indices in a list of prompts.
        
        This function is very important. It can be used before calling other functions
        (logit_lens, attention_map_generation, run_patching) so it will be clear what 
        is the position of each token. Understanding token positions is critical for 
        accurate interpretation of model internals and for targeted interventions.
        
        Args:
            prompt: Input text, list of prompts (List[str])
            
        Returns:
            Formatted string with numbered prompts and their tokens
            
        Example:
            >>> # List of prompts
            >>> prompts = ["Hello world", "Hi there"]
            >>> model_invest.get_token_indices_in_prompt(prompts)
            Prompt 1: "Hello world"
            Tokens: [('Hello', 0), (' world', 1)]
            
            Prompt 2: "Hi there"
            Tokens: [('Hi', 0), (' there', 1)]
            
            >>> # For character-level tokenizers, each character might be a token
            >>> prompt = "Hi!"
            >>> model_invest.get_token_indices_in_prompt(prompt)
            [('H', 0), ('i', 1), ('!', 2)]
        """
            pass
    
    
    def attention_map_generation(self, prompts: List[str], query_positions: List[int], layer_head_pairs: List[Tuple[int, int]]) -> str:
        """
        Generate attention maps for multiple prompts across multiple layer-head combinations.
        
        Args:
            prompts: List of input texts
            query_positions: List of positions in prompts to examine attention from (one per prompt)
            layer_head_pairs: List of tuples (layer, head) to examine

        Returns:
            String description of attention maps for all combinations
            
        Example:
            
            >>> # List of prompts and layer-head pairs
            >>> prompts = ["Hello world", "The cat sat"]
            >>> query_positions = [0, 1]  # First token for prompt 1, second token for prompt 2
            >>> layer_head_pairs = [(2, 1), (3, 5), (4, 0)]
            >>> attention_maps = model_invest.attention_map_generation(prompts, query_positions, layer_head_pairs)
            
        """
        pass

        
    def run_patching(self, source_prompts: List[str], counterfactual_prompts: List[str], 
                Optional[int]]], token_positions_source: List[int], 
                token_positions_counterfactual: List[int], layer_head_pairs: List[Tuple[int, tokens: List[str] = None, top_k: int = 10) -> str:
        """
   
   Performs causal intervention by replacing activations from source prompts with 
   counterfactual activations at specified positions. Analyzes probability changes 
   to determine whether components store semantic content vs. structural patterns.
   
   Args:
       source_prompts: List of clean/original prompts
       counterfactual_prompts: List of prompts with alternative entities/contexts
       token_positions_source: Token positions in source prompts to patch FROM
       token_positions_counterfactual: Token positions in counterfactual prompts to patch TO
       layer_head_pairs: List of (layer, head) tuples. Use None for head to patch MLP
       tokens: Specific tokens to analyze probability changes for (optional)
       top_k: Number of top probability tokens to return if tokens not specified
       
   Returns:
       Formatted results showing probability changes for each layer-head combination
       
   Example:
       >>> # Test what "Italy" activation encodes
       >>> sources = ["The capital city of France is Paris. The capital city of Italy is"]
       >>> counterfactuals = ["The capital city of Austria is Vienna. The capital city of Germany is"]  
       >>> layer_head_pairs = [(8, 3), (10, 5), (12, None)]  # Test attention heads + MLP
       >>> results = model_invest.run_patching(sources, counterfactuals,
       ...                                   token_positions_source=[10],  # "Italy" position
       ...                                   token_positions_counterfactual=[10], # "Germany" position  
       ...                                   layer_head_pairs)
   
   
   **Experimental Guidelines:**
   
   1. **Prompt Design**: Source prompts must stop right before the key prediction token.
      The probability shift of this key token reveals what the patched activation encoded.
      Examples: "The capital city of France is", "The author of 1984 is", "I want to eat"
   
   2. **What Properties to Test**:
      
      **Value Representation Test:**
      - Patch entity A -> entity B 
      - If probability increases for entity B's associated answer: activation stores semantic identity
      - Example: Italy->Austria patch increases "Vienna" probability = country identity encoded
      
      **Position Representation Test:**
      - Same patch as above
      - If probability increases for entity A's associated answer: activation stores structural position  
      - Example: Italy->Austria patch increases "Paris" probability = positional pattern encoded

      
   **Interpretation Framework:**
   
   **Direction of Shift Reveals Representational Content Type:**
   
    ### PATCHING INCREASES CORRECT TOKEN PROBABILITY (patched_probabilities - clean_probabilities > 0):
    **Interpretation**: The original component was INTERFERING with correct prediction
    **Action**: Removing/noising this component would IMPROVE performance
    **Example**: Patch increases "Rome" probability -> head was suppressing the correct answer

    ### PATCHING DECREASES CORRECT TOKEN PROBABILITY (patched_probabilities - clean_probabilities < 0):  
    **Interpretation**: The original component was CONTRIBUTING POSITIVELY to prediction
    **Action**: This component is NECESSARY for optimal performance
    **Example**: Patch reduces "Rome" probability -> head was helping promote the correct answer

    ###PATCHING DOESN'T AFFECT CORRECT TOKEN PROBABILITY (patched_probabilities - clean_probabilities == 0):  
    **Interpretation**: 
    Option 1: The component has no causal influence on the target token prediction
    Option 2: Both clean and corrupted activations encode the same information relevant to this prediction
    **Action**: To distinguish between these scenarios, test multiple counterfactuals that encode different types of information


    **Pro Tip**: Always test multiple entity pairs and prompt variations to ensure 
    findings generalize beyond specific examples.

        """

        pass

    def logit_lens(self,
       prompts: List[str],
       token_positions: List[int],
       layer_head_pairs: List[Tuple[int, Optional[int]]],
       tokens: List[str] = None,
       top_k: int = 20) -> str:
    """
        Apply logit lens to examine internal predictions across multiple layer-head combinations.
        
        Args:
            prompts: List of input text prompts
            token_positions: Positions in prompts to examine (one per prompt)
            layer_head_pairs: List of tuples (layer, head) to examine. Use None for head to examine MLP output
            tokens: List of specific tokens to analyze (optional)
            top_k: Number of top tokens to return if tokens not specified
            
        Returns:
            String description of logit lens results for all layer-head combinations
            
        Example:
            >>> # Multiple prompt analysis across multiple layer-head pairs
            >>> prompts = ["The Eiffel Tower is located in", "The capital of France is"]
            >>> layer_head_pairs = [(5, None), (8, 4), (10, 7)]  # Mix of MLP and attention heads
            >>> token_positions = [-1, -1]  # Last token for both prompts
            >>> results = model_invest.logit_lens(prompts, token_positions, layer_head_pairs)
     
            >>> # Analyze specific tokens
            >>> tokens_to_check = ["Paris", "London", "New York", "Rome", "Tokyo"]
            >>> results = model_invest.logit_lens(prompts, token_positions, tokens=tokens_to_check, layer_head_pairs)
        """
        pass
\end{pythoncode}
\label{fig:prompt}

\subsection{Agent's system prompt}
\label{app:agent_system_prompt}
The agent’s system prompt is structured into the following sections:
\begin{enumerate}
\item A task description specifying the component under analysis (e.g., Layer~6, Head~8), including task examples and the token position of interest.
\item Descriptions of the available interpretability tools.
\item General guidance on interpreting experimental results.
\item Instructions for formatting intermediate analyses, tool calls, and the final hypothesis.
\item Initial results from logit lens and attention pattern analyses.
\end{enumerate}

\begin{pythoncode}[title=System Prompt, breakable]

You are a mechanistic interpretability agent. Your task is to analyze specific components of a language model. Your goal is to understand and describe the role or behavior of these components within the context of provided prompts and token positions, ultimately forming a unified hypothesis that explains the component's function across the entire dataset.

Here is the information you'll be working with:

<user_guidelines>
{user_guidelines}
</user_guidelines>

<prompts_and_positions>
{format_prompts_and_positions(model, prompts, positions)}
</prompts_and_positions>

<component_info>
{component}
</component_info>

You have access to an API for a mechanistic interpretability toolkit. Here are the available functions:

<api_functions>
{RESEARCHER_API}
</api_functions>

**Function Call Syntax:**
When calling API functions that require prompt or position information, use the following syntax:
- `prompts[i]` to reference the i-th prompt in the dataset
- `token_positions[i]` to reference the token positions for the i-th prompt
- Replace `i` with the actual index number (0, 1, 2, etc.)

Example function calls:
- `get_token_indices_in_prompt(prompts[0])` for the first prompt
- `logit_lens(prompts[1], token_positions[1])` for the second prompt
- `attention_map_generation(prompts[2], token_positions[2])` for the third prompt

Analysis Process:
1. Form initial hypotheses about the component's role based on its type and potential interactions with other components. Pay attention to positional dependencies and how the component's behavior might vary based on token positions within the prompt.

2. Design and run experiments using the API functions to test your hypotheses across ALL prompts simultaneously. Design experimants that refine or falsify your hypotheses. For each experiment, state its purpose and what you hope to learn from it about the tested hypothesis. Run experiments on both the given prompts and on new prompts that are not part of your original dataset to test and validate your hypotheses.

3. CRITICAL - Run Patching Experiments: After initial observations, you MUST run patching experiments using the `run_patching` function to test causal relationships. Patching experiments are essential for:
   - Verifying whether the component is causally responsible for observed behaviors
   - Testing if ablating or modifying the component's output changes model's prediction according to the hypothesis
   - Distinguishing correlation from causation in component behavior
   - Validating hypotheses about the component's functional role
   - Validating hypotheses about stored positional information

4. Analyze the results of these experiments and summarize key findings across the entire dataset. Look for patterns in how the component's function depends on position - does it behave differently at the beginning, middle, or end of sequences? Does it show sensitivity to relative positions between tokens?

5. Every time you get new results, come up with 2-5 hypotheses that could explain what you've observed so far inside <hypothesis_i> tags. For each hypothesis, design a set of experiments that will help you determine which one is most plausible. Incorporate positional dependencies into your hypotheses - explain not just what the component does, but how its function varies with position. Use patching results to confirm or falsify your causal claims.

6. CRITICAL: Your ultimate goal is to develop a final unified hypothesis that explains the component's role across the majority of prompts in the dataset, not just individual cases. 

When planning experiments, use the following format:


<experiment_calls>
<description>
[Purpose of the experiment]
</description>
<experiment>
[API function call]
</experiment>
[Repeat for each experiment]
</experiment_calls>

For example: 

<experiments_calls>
<description>
Test if position matters - what if the second animal comes first in the repeated pattern?
</description>
<experiment>
mi.attention_map_generation([prompts[0], prompts[1], prompts[2], prompts[3], prompts[4]], query_positions=[53, 53, 53, 53, 53], layer_head_pairs=[(15,13), (2, None)])
</experiment>
<description>
Test what happens with names that appear more than twice
</description>
<experiment>
mi.logit_lens([prompts[0], prompts[1], prompts[2], prompts[3], prompts[4]],token_positions=[53, 53, 53, 53, 53], layer_head_pairs=[(15, 13)])
</experiment>
</experiments_calls>

Assume there is an active instance mi of ModelInvest when making these calls. VERY IMPORTNAT - use <experiment_calls> tags!.

After receiving experiment results (which will be provided in <experiments_results> tags), analyze them and update your hypotheses. Wrap your thought process in <thought_process> tags:


<thought_process>
  <analysis>
    [Insert analysis here]
  </analysis>

  <hypotheses>
    <hypothesis_1>
      [First hypothesis]
    </hypothesis_1>

    <hypothesis_2>
      [Second hypothesis]
    </hypothesis_2>

    <hypothesis_3>
      [Third hypothesis]
    </hypothesis_3>

    <hypothesis_4>
      [Fourth hypothesis]
    </hypothesis_4>

    <hypothesis_5>
      [Fifth hypothesis]
    </hypothesis_5>
  </hypotheses>
</thought_process>


If you think your hypothesis is proved and you don't need to run any more experiments, instead of <experiment_calls>, write your final conclusion in <final_hypothesis> tags.


**Important Guidelines:**
- Only base your hypotheses and conclusions on actual experiment results.
- Avoid making unsupported assumptions.
- Communicate clearly within the specified tags.
- PATCHING EXPERIMENTS ARE MANDATORY: You must run patching experiments to test causal relationships. Observational experiments alone are insufficient for establishing the component's true functional role.
- CROSS-PROMPT ANALYSIS: Always analyze patterns and behaviors across ALL prompts to identify the component's general function, not just prompt-specific behaviors.
- UNIFIED HYPOTHESIS REQUIREMENT: Your final hypothesis must explain the component's role in a way that is relevant and applicable to the majority of prompts in the dataset, including any systematic positional dependencies.
- Continue the process of analyzing, hypothesizing, experimenting, and responding to feedback until you have a clear understanding of the component's role or behavior across the entire dataset.
- Use proper indexing: When calling functions, you can use `prompts[i]` and `token_positions[i]` where i corresponds to the specific prompt index you want to analyze. 
- When runing patching experimants, it is better to come up with a counterfactual prompt, than to use the provided prompts.
- Only use experiment results that you receive from the user. Don't generate experiment results yourself.
- If you believe your unified hypothesis is proven across most prompts, state your final conclusion in the <final_hypothesis> tags.
- Run experiments on both the given prompts and on new prompts that are not part of your original dataset to test and validate your hypotheses.
- Run patching experiments to test whether this component is passing semantic or positional information.

**Experimental Strategy:**
- Design experiments that test the component's behavior across multiple prompts simultaneously using different indices
- Run patching experiments systematically: 
  - Start with observational experiments (attention maps, logit lens) to form hypotheses
  - Follow up with patching experiments to test causal relationships
  - To analyze patching effects: If patching the component **DECREASES** token probability, the component was contributing positively to that token's prediction in the original forward pass. If patching the component **INCREASES* token probability, the component was interfering with or suppressing that token's prediction in the original forward pass.
  - When analyzing the patching experiments, take into account the information that you patched. It might contain semantic information or positional information, or it might be a non-relevant activation that fully ablte the head. 
  - Run patching experiments to test whether this component is passing semantic or positional information.
  - Make sure to run the patching experiments at the given token positions. Patching at a different token position is not guaranteed to have a causal effect.
  - Design counterfactual prompts that help distinguish between positional and semantic information.
  - token_position_source and token_position_counterfactual do not need to be the same.
- Run attention analysis:
    - Distinguish between heads that respond to positional information (relative positioning, absolute indices) versus semantic content (syntactic roles, entity types, conceptual relationships)
    - Test your hypothesis on counterfactual prompts
    - Test each head's sensitivity to positional vs. semantic information
    - Examine consistency patterns across different prompts by analyzing:
        - **Segment Consistency**: Whether attention targets the same content segments across prompts
        - **Position Consistency**: Whether attention focuses on the same token positions or at the same relative token positions
- Run logit lens analysis:
    - Analyze what the output component encodes:
        - Identify tokens directly referencing input prompt elements
        - Find semantically related tokens that aid inference
        - Detect conceptual links beyond explicit prompt content
    - Recognize that tokens may be subwords (prefixes, suffixes, fragments). Determine if these subword patterns relate to the encoded information above.

**Positional Information to Consider:**
- Absolute position: Token index within the sequence
- Relative position: Distance between tokens of interest 


**Final Hypothesis Requirements:**
Your final hypothesis should:
1. Explain the component's primary role/function in general terms
2. Be supported by evidence from experiments across multiple prompts and positions
3. Include evidence from patching experiments that demonstrate causal relationships, from logit lens experiments, and from attention pattern analysis.
4. Account for the majority of observed behaviors in the dataset
5. Address any significant exceptions or outliers
6. Clearly distinguish between correlational observations and causally verified mechanisms
7. Be concise yet comprehensive enough to capture the component's essential function
8. Contains summarized descriptio inside <summarized_description> tags, 1-3 sentences that describe your findings.

Begin your analysis by examining all prompts to identify common patterns, positional structures, and semantic relationships, then form initial hypotheses about the component's general role based on its type, position, semantic context, and potential interactions with other components across the entire dataset. 


{initial_results}

Be concise in your answers!


\end{pythoncode}

\subsection{Clustering}
\label{ap:cluatering}
Instructions for clustering components based on their functional descriptions.
\begin{pythoncode}[title=Clustering Prompt, breakable]

You are an AI assistant with expertise in mechanistic interpretability. Your task is to analyze experimental results on the role of language model components in a certain task and at specific token positions, group the heads into clusters with similar functions, and provide a brief description of each cluster's shared function.

Prompts and Positions:
<prompts>
{format_prompts_and_positions(model, prompts, positions)}
</prompts>

Here are the experiment results you will be analyzing:
<experiment_results>
{experiment_results}
</experiment_results>

Please follow these steps to complete the task:
1. Carefully review the experiment results provided above.
2. Identify patterns and similarities in the function of different components.
3. Group the heads into clusters based on their shared functions.
4. For each cluster you identify, provide the following information:
   a. List of heads in the cluster (by layer and head number)
   b. A brief description of the shared function

5. Present your analysis in the following format:
<cluster_analysis>
<cluster>
<cluster_name>[indicative short name]</cluster_name>
<components>
<component>layer X head Y</component>
<component>layer X head Y</component>
<component>mlp Y</component>
<component>mlp Y</component>
Repeat for each identified component]
</components>
<function>[Brief description of shared function. 1-3 sentences]</function>
<evidence>[Summary of experimental evidence]</evidence>
</cluster>
[Repeat for each identified cluster]
</cluster_analysis>

Remember to keep your descriptions clear and concise, focusing on the key aspects of each cluster's function and activation pattern. Avoid speculation and stick to the information provided in the experiment results.

If you need to organize your thoughts before providing the final analysis, you may use a <scratchpad> section to brainstorm and outline your ideas.

Remember: YOU HAVE to assign each of the components into a cluster!

Begin your analysis now.
\end{pythoncode}

\newpage

\section{Tasks}
\label{ap:tasks}
\subsection{Indirect Object Identification (IOI) - GPT-2-Small}
\label{ap:task-ioi}
The IOI task \citep{wang2022interpretability} is one of the earliest and most well-known examples of circuit analysis. Prompts in this task follow the form "When Mary and John went to the store, John gave a drink to", where the model is expected to predict "Mary". Table~\ref{tab:ioi-head-classes} shows the attention heads grouped into clusters along with their associated functionalities.

\begin{table*}[h]
\caption{Attention heads involved in the IOI task in GPT-2-small model, grouped according to their functional roles as reported in the original paper.}
\centering
\footnotesize
\setlength{\tabcolsep}{6pt}
\renewcommand{\arraystretch}{1.15}
\resizebox{\textwidth}{!}{%
\begin{tabular}{l l l}
\toprule
\textbf{Cluster Name} & \textbf{Components (Layer, Head)} & \textbf{Description} \\
\midrule

Duplicate Token Heads
& \parbox{3cm}{(0,1), (0,10), (3,0)}
& \parbox{9cm}{These heads detect when the current token is a duplicate of one of the previous tokens in the sequence. Specifically, they are active at the second occurrence of the subject (actor) and attend back to the first occurrence of the subject. When they detect this duplication, they write positional information about the duplicate token's location into the residual stream. This positional signal is later used by S-Inhibition Heads to help Name Mover Heads avoid attending to duplicated tokens.} \\
\addlinespace

S-Inhibition Heads
& \parbox{3cm}{(7,3), (7,9), (8,6), (8,10)}
& \parbox{9cm}{These heads attend to the second occurrence of the subject token (the actor) and influence the Name Mover Heads' attention. This inhibits the Name Mover Heads' attention to the first and second occurrences of the subject tokens (the actor's name), effectively removing duplicated tokens from consideration. They achieve this through a combination of token signals and position signals.} \\
\addlinespace

Name Mover Heads
& \parbox{3cm}{(9,6), (9,9), (10,0)}
& \parbox{9cm}{These heads are responsible for identifying and outputting the indirect object (the recipient). They attend to the recipient's name in the sentence and copy it forward. Due to the influence of S-Inhibition Heads, they preferentially attend to the indirect object token rather than the subject token(s).} \\
\addlinespace

Previous Token Heads
& \parbox{3cm}{(2,2), (4,11)}
& \parbox{9cm}{These heads copy information from the immediately previous token to the current position. Specifically, they copy information about the first-occurrence subject token (the actor) to the position right after it. This copied information enables Induction Heads to recognize the pattern when the repeated subject token appears later.} \\
\addlinespace

Induction Heads
& \parbox{3cm}{(5,5), (5,8), (5,9), (6,9)}
& \parbox{9cm}{These heads attend to the token after the first occurrence of the subject (mediated by Previous Token Heads). They recognize patterns of the form [A][B]...[A] and contribute to predicting [B]. In this task, they provide signals that the subject is duplicated and indicate the position of its first occurrence.} \\
\addlinespace

Negative Name Mover Heads
& \parbox{3cm}{(10,7), (11,10)}
& \parbox{9cm}{These heads write in the opposite direction of Name Mover Heads, decreasing confidence in predictions. They share similar properties with Name Mover Heads, except that they write opposite to the names they attend to; one hypothesis is that they help hedge against overconfident errors.} \\
\bottomrule
\end{tabular}%
}

\label{tab:ioi-head-classes}
\end{table*}

\newpage
\subsection{Indirect Object Identification (IOI) - Pythia-160M}

\citet{DBLP:conf/nips/Tigges0YB24} has analyzed the IOI circuit in Pythia-160M. \citep{biderman2023pythia}. Table~\ref{tab:ioi-pythia-head-classes} shows the attention heads grouped into clusters along with their associated functionalities.

\begin{table*}[h]
\caption{Attention heads involved in the IOI task in Pythia-160M, grouped according to their functional roles as reported in the original paper.}
\centering
\footnotesize
\setlength{\tabcolsep}{6pt}
\renewcommand{\arraystretch}{1.15}
\resizebox{\textwidth}{!}{%
\begin{tabular}{l l l}
\toprule
\textbf{Cluster Name} & \textbf{Components (Layer, Head)} & \textbf{Description} \\
\midrule

S-Inhibition Heads
& \parbox{3cm}{(6,6), (7,2), (7,9)}
& \parbox{9cm}{These heads attend to the second occurrence of the subject token (the actor) and suppress attention to duplicated subject tokens. They influence downstream Name Mover Heads by inhibiting attention to both the first and second occurrences of the subject, effectively removing subject names from consideration when predicting the indirect object. Their behavior combines token-identity signals with positional information.}  \\
\addlinespace
\hline
% \addlinespace
Name Mover Heads
& \parbox{3cm}{(8,10), (8,2), (9,4), (10,7)}
& \parbox{9cm}{These heads are responsible for identifying and copying the indirect object (the recipient's name) to the output. They attend strongly to the recipient token and write its identity into the residual stream at the final position. Due to suppression signals from S-Inhibition Heads, they preferentially attend to the indirect object rather than to duplicated subject tokens.} \\ 
\addlinespace
\hline
% \addlinespace
Previous Token Heads
& \parbox{3cm}{(2,6)}
& \parbox{9cm}{This head copies information from the immediately preceding token to the current position. In the IOI task, it propagates information about the first occurrence of the subject token to the following position. This copied signal enables Induction Heads to later recognize repeated subject patterns and is a key component of the induction mechanism.} \\
\addlinespace
\hline
% \addlinespace
Induction Heads
& \parbox{3cm}{(4,11), (4,6), (5,0)}
& \parbox{9cm}{These heads attend to the token following the first occurrence of the subject, mediated by signals from Previous Token Heads. They detect repeated patterns of the form [A][B]$\ldots$[A] and contribute to predicting [B]. In the IOI task, they signal both that the subject is duplicated and the position of its first occurrence.} \\
\addlinespace
\hline
% \addlinespace
Negative Name Mover Heads
& \parbox{3cm}{(9,1)}
& \parbox{9cm}{This head shares attention patterns with Name Mover Heads but writes in the opposite direction, decreasing the logits of the names it attends to. It is hypothesized to act as a corrective mechanism, reducing overconfident predictions and helping the model hedge when uncertain.} \\
\addlinespace
\hline
% \addlinespace
Negative Copy Suppression Heads
& \parbox{3cm}{(9,5)}
& \parbox{9cm}{These heads are active at the final token position in the sentence, where the model predicts the output. They attend to earlier tokens in the input and reduce the model's confidence in those tokens when they are strongly represented in the residual stream. In the IOI circuit, they reduce the model’s confidence in the correct indirect object name, which harms task performance as they push the model toward incorrect outputs.} \\
\addlinespace
\hline
% \addlinespace
Positive Copy Suppression Heads
& \parbox{3cm}{(8,9)}
& \parbox{9cm}{These heads are active at the final token position in the sentence, where the model predicts the output. They attend to earlier tokens in the input and reduce the model's confidence in those tokens when they are strongly represented in the residual stream. In the IOI circuit, these heads downweight the incorrect subject name. By suppressing the wrong name, these heads allow the correct indirect object to dominate the output logits, thereby improving task performance.}  \\
\bottomrule
\end{tabular}%
}
\label{tab:ioi-pythia-head-classes}
\end{table*}

\newpage
\subsection{Greater-Than}
The Greater-Than task \citep{hanna2023does} studies numerical comparison in GPT-2 small using prompts such as "The war lasted from the year 1732 to the year 17", where the model should predict end years greater than 32. We adopt the same prompts as the original work. Table~\ref{tab:gt-head-classes} shows attention heads grouped by functionality.

\begin{table*}[h]
\centering
\footnotesize
\setlength{\tabcolsep}{6pt}
\renewcommand{\arraystretch}{1.15}
\resizebox{\textwidth}{!}{%
\begin{tabular}{l l l}
\toprule
\textbf{Cluster Name} & \textbf{Components (Layer, Head)} & \textbf{Description} \\
\midrule

Greater-Than Computer
& \parbox{5cm}{MLP8, MLP9, MLP10, MLP11}
& \parbox{9cm}{These components implement the mathematical comparison operation. They are active at the final token position where the model predict the end year. They directly boost the probabilities for all years greater than the start year YY while suppressing probabilities for years less than or equal to YY.} \\ \\
% \addlinespace

Year Identification &
\parbox{5cm}{(5,1), (5,5), (6,1), (6,9), (7,10), (8,8), (8,11), (9,1)}
& \parbox{9cm}{These components are responsible for detecting and extracting the start year value from the input sentence. They are active at the final token position (where the model predicts the end year) and attend back to the YY position (the two-digit start year token) in the template ""The <noun> lasted from the year XXYY to the year XX"". Their output corresponds to the start year YY value in the embedding space, transferring it to the greater-than computer, effectively communicating ""this is the reference number"" to the downstream MLPs. Their attention patterns show consistent focus on the YY token regardless of the specific year value, indicating they've learned to identify this positionally and contextually important numerical information.} \\ 
\bottomrule
\end{tabular}%
}
\caption{Attention heads involved in the Greater-Than task in GPT-2-small model, grouped according to their functional roles as reported in the original paper.}
\label{tab:gt-head-classes}
\end{table*}

\newpage
\subsection{Acronyms}
The Acronym Prediction task \citep{DBLP:conf/aistats/Garcia-Carrasco24} examines how GPT-2 small predicts three-letter acronyms from their expanded forms. Prompts follow the form "The Chief Executive Officer $->$ CE", where the model is expected to predict "O". Table~\ref{tab:ac-head-classes} shows the attention heads grouped into clusters along with their associated functionalities.

\begin{table*}[h]
\centering
\footnotesize
\setlength{\tabcolsep}{6pt}
\renewcommand{\arraystretch}{1.15}
\resizebox{\textwidth}{!}{%
\begin{tabular}{l l l}
\toprule
\textbf{Cluster Name} & \textbf{Components (Layer, Head)} & \textbf{Description} \\
\midrule

Letter Mover Heads
& \parbox{3cm}{(8,11), (10,10), (9,9), (11,4)}
& \parbox{9cm}{These heads are responsible for the core acronym formation process. They are active at the position where the previous acronym letter was predicted (for example, when the model has already predicted "CE" and needs to predict the third letter "O"). These heads attend back to the capital letter at the beginning of the corresponding word (like attending to "O" in "Officer" when predicting "O" in the acronym). They perform a copying mechanism: they take the capital letter of the corresponding word and copy it to form the predicted capital letters for the acronym.} \\ \\
% \addlinespace

Fuzzy Previous Token Heads
& \parbox{3cm}{(4,11), (1,0), (2,2)}
& \parbox{9cm}{These attention heads handle information flow between words in the input phrase. They are active at each word's capital letter position and attend to the previous word's capital letter. These heads follow a "previous token" pattern, systematically copying information from one word to the next in sequence. These heads create a chain of information that flows from word to word, ensuring that context from e.g. "Executive" is available when processing "Officer," and so on. This information is then moved to the Letter Mover Heads through the Information Bridge Heads to ensure that when the model needs to predict the third letter "O" when given "CE," it can access information not just about "Executive" but also about its relationship to the next word "Officer".} \\ \\

Information Bridge Heads
& \parbox{3cm}{(5,8)}
& \parbox{9cm}{These attention heads move word-level information to the acronym-level predictions. They are active at acronym letter positions and attend to both the corresponding word's capital letter and surrounding word content. They connect the Fuzzy Previous Token Heads with the Letter Mover Heads, moving information about the letters for the acronym.} \\
\bottomrule
\end{tabular}%
}
\caption{Attention heads involved in the Acronyms task in GPT-2-small model, grouped according to their functional roles as reported in the original paper.}
\label{tab:ac-head-classes}
\end{table*}

\newpage
\subsection{Colored-Objects}

The Colored Objects task \citep{merullo2023circuit} requires the model to generate the color of an object that was previously described in context, among other objects. Prompts follow a one-shot format: "Q: On the table, there is a blue pencil, a black necklace, and a yellow lighter. What color is the pencil? A:", where the model is expected to predict "blue". Table~\ref{tab:co-head-classes} shows the attention heads grouped into clusters along with their associated functionalities.

\begin{table*}[h]
\centering
\footnotesize
\setlength{\tabcolsep}{6pt}
\renewcommand{\arraystretch}{1.15}
\resizebox{\textwidth}{!}{%
\begin{tabular}{l l l}
\toprule
\textbf{Cluster Name} & \textbf{Components (Layer, Head)} & \textbf{Description} \\
\midrule

Mover Heads
& \parbox{3cm}{(10,6), (14,5), (14,14), (15,10), (15,14), (16,0), (16,15), (17,4), (18,12), (19,15), (21,7)}
& \parbox{9cm}{These heads are responsible for copying the correct answer token to the next prediction. They are active at the final token position (the ":" token) and attend back to the correct color token in the context. These heads write strongly in the embedding direction of whatever they attend to, effectively telling the model to predict that token.} \\ \\
\addlinespace

Content Gatherer Heads
& \parbox{3cm}{(11,6), (11,7), (12,6), (12,14), (12,15), (13,1), (13,3)}
& \parbox{9cm}{These heads tell the mover heads which color to select. They are active at the final token position and attend primarily to the queried object and the token "color" in the question. These heads gather information about which specific object's color is being requested and pass this signal to guide the mover heads' attention.} \\ \\
\addlinespace

Induction/Duplicate Token Heads
& \parbox{3cm}{(6,4), (6,15), (7,2), (7,4), (7,11), (8,15), (9,3), (10,12), (10,14)}
& \parbox{9cm}{These heads detect the duplication pattern where the same object appears twice in the prompt---once in the object list and again in the question. They are active at the second occurrence of the object (in the question) and attend back to the first occurrence or the token after it. This duplication detection provides a signal that gets passed to content gatherer heads, helping the model to locate the queried object in the object list.} \\ \\
\bottomrule
\end{tabular}%
}
\caption{Attention heads involved in the Colored Objects task in GPT-2-XL model, grouped according to their functional roles.}
\label{tab:co-head-classes}
\end{table*}
\newpage
\subsection{Entity-Tracking}

The Entity Tracking task \citep{prakash2024fine} evaluates a model's ability to track entities and their associated properties across context. Prompts follow the form "The apple is in Box F, the computer is in Box Q, the document is in Box X... Box F contains the", where the model is expected to predict "apple". Table~\ref{tab:et-head-classes} shows the attention heads grouped into clusters along with their associated functionalities.

\begin{table*}[h]
\centering
\resizebox{\columnwidth}{!}{%
\begin{tabular}{l l l}
\toprule
\textbf{Cluster Name} & \textbf{Components (Layer, Head)} & \textbf{Description} \\
\midrule

Value Fetcher Heads
& \parbox{7cm}{(14,13), (15,2), (15,5), (15,6), (15,13), (15,18), (16,23), (16,27), (17,3), (17,27), (17,28), (18,3), (18,6), (18,8), (18,20), (18,25), (18,28), (19,8), (19,11), (19,20), (19,26), (19,30), (20,14), (20,29), (21,0), (21,3), (21,4), (21,25), (21,26), (23,5), (23,15), (23,16), (23,17), (23,19), (23,20), (24,0), (24,5), (24,8), (25,14), (29,7)}
& \parbox{9cm}{These heads are responsible for identifying and outputting the correct object (the entity being tracked). They attend to object tokens in the sentence corresponding to the queried box. They then copy the value of the object they attend to. Thanks to the contribution of the Position Transmitter Heads, which provide positional guidance through query composition, they preferentially attend to the object associated with the queried box rather than other objects in the context.} \\ \\
\addlinespace

Position Transmitter Heads
& \parbox{7cm}{(11,23), (12,23), (14,27), (16,2), (19,12)}
& \parbox{9cm}{These heads write to the query vectors of the Value Fetcher Heads. They attend to the box label in the query, and transport positional information from the Position Detector Heads to guide the Value Fetcher Heads toward the correct object location.} \\ \\
\addlinespace

Position Detector Heads
& \parbox{7cm}{(1,9), (4,4), (6,10), (7,9), (7,17), (7,26), (8,22), (9,2), (9,7), (9,10), (9,21), (10,3), (10,4), (11,23)}
& \parbox{9cm}{These heads identify positional information of the correct object within the context. They attend to the correct box label, and receive contextual structure information from Structure Reader Heads through value composition and pass their positional information to Position Transmitter Heads.} \\ \\
\addlinespace

Structure Reader Heads
& \parbox{7cm}{(8,12), (8,21), (11,9), (11,23), (12,23)}
& \parbox{9cm}{These heads process the contextual structure and gather information related to each box separately. They attend primarily to tokens within a sentence describing a single box. They collect information about the overall sentence structure. Their output provides contextual understanding to Position Detector Heads through value composition.} \\
\bottomrule
\end{tabular}
}
\caption{Attention heads involved in the Entity Tracking task in LLaMa-7B model, grouped according to their functional roles.}
\label{tab:et-head-classes}
\end{table*}
\newpage

% === PREAMBLE: Add to your document preamble ===
% \usepackage[most]{tcolorbox}
% \tcbuselibrary{listings, breakable, skins}
% \usepackage{listings}
% \usepackage{array}

% Box definitions
\newtcolorbox{systempromptbox}[1][]{
  colback=gray!5,
  colframe=gray!70,
  fonttitle=\bfseries,
  title=System Prompt,
  breakable,
  left=2mm, right=2mm,
  #1
}

\newtcolorbox{taskbox}[1][]{
  colback=blue!3,
  colframe=blue!40,
  fonttitle=\bfseries,
  title=Task Setup,
  breakable,
  left=2mm, right=2mm,
  #1
}

\newtcolorbox{usermsg}[1][]{
  colback=blue!5,
  colframe=blue!50,
  fonttitle=\bfseries,
  title=User,
  breakable,
  left=2mm, right=2mm,
  #1
}

\newtcolorbox{agentmsg}[1][]{
  colback=blue!5,
  colframe=blue!50,
  fonttitle=\bfseries,
  title=Agent,
  breakable,
  left=2mm, right=2mm,
  #1
}

\newtcolorbox{resultbox}[1][]{
  colback=yellow!5,
  colframe=orange!50,
  fonttitle=\bfseries,
  title=Experiment Results,
  breakable,
  left=2mm, right=2mm,
  #1
}

\newtcolorbox{hypothesisbox}[1][]{
  colback=purple!5,
  colframe=purple!50,
  fonttitle=\bfseries,
  title=Final Hypothesis,
  breakable,
  left=2mm, right=2mm,
  #1
}

\lstset{
  basicstyle=\ttfamily\small,
  breaklines=true,
  breakatwhitespace=false,
  columns=flexible,
  keepspaces=true,
  showstringspaces=false,
  frame=none,
  xleftmargin=0pt
}

% === END PREAMBLE ===

\section{Example Agent Run}
\label{appendix:agent-conversation}

An example of the agent's output during analysis and final hypothesis generation. This trace is from a run analyzing head (9,9) on the IOI task, labeled a "name mover head" in the original paper, which is responsible for passing the recipient's name to the final token position.

\subsection{Agent Iteration}
\label{ap:agent-iter}
Each iteration of the agent consists of three stages: (1) analyzing the results just received, (2) proposing multiple hypotheses for the component's functional role, and (3) designing new experiments, with a stated purpose for each.

\begin{agentmsg}
\begin{minipage}{1\textwidth}

\textbf{Analysis:}

Examining the initial results for Layer 9, Head 9:

1. **Attention Pattern Analysis**: Across all prompts, when querying from the final "to" position, the attention head shows a strong, consistent pattern - it attends predominantly to the first person mentioned who is NOT the giver (i.e., the recipient). For example:
   - Prompt 1: Attends to "Joseph" (0.81) when Gregory is giving
   - Prompt 2: Attends to "Ryan" (0.87) when Kenneth is giving
   - Prompt 3: Attends to "Stephanie" (0.87) when Joshua is giving

2. **Logit Lens Results**: The head's output strongly promotes the name it's attending to as the next token prediction. The probabilities are remarkably high (often 0.9+ for the attended name).

3. **Position Pattern**: The head operates at the final "to" position and looks back to identify the non-giver person from the pair introduced earlier.

\textbf{Hypotheses:}
\begin{enumerate}
    \item Layer 9, Head 9 is a "recipient identification" head that identifies and promotes the non-giver person from a pair as the recipient of an action when positioned at "to" following a giving verb phrase.
    \item The head encodes a more general "other person" retrieval mechanism - it identifies and promotes the person who is not currently performing the action, regardless of the specific verb or context.
    \item The head specifically tracks the first-mentioned person in "X and Y" constructions and promotes them when they're not the subject of the current clause, implementing a form of discourse tracking.
    \item The head implements a simple positional heuristic: attend to names at specific relative positions (early in the sequence) and promote them as likely continuations at the end of giving-related phrases.
    \item The head combines semantic understanding (giving actions) with syntactic patterns (prepositional phrases ending in "to") to identify and promote the most likely indirect object from previously mentioned entities.
\end{enumerate}

\textbf{Experiments:}

\textbf{Purpose:} Test if the head's behavior is specific to giving actions or works with other verb types by creating prompts with different action types

\begin{lstlisting}[language=Python]
mi.attention_map_generation(["Sarah and Michael went to the park. Sarah talked to", "Emma and David were at home. Emma listened to", "Robert and Lisa entered the room. Robert looked at"], query_positions=[10, 10, 10], layer_head_pairs=[(9, 9)])
\end{lstlisting}

\textbf{Purpose:} Test the causal role of this head by patching - replace the head's activation when it's attending to person A with activation from a prompt where it attends to person B

\begin{lstlisting}[language=Python]
mi.run_patching(source_prompts=["Alex and Maria went to school. Alex gave a book to"], 
                counterfactual_prompts=["John and Sarah went to school. Sarah gave a book to"],
                token_positions_source=[13],
                token_positions_counterfactual=[13], 
                layer_head_pairs=[(9, 9)],
                tokens=["Maria", "Alex", "Sarah", "John"],
                top_k=5)
\end{lstlisting}

\end{minipage}
\newpage
\begin{minipage}{1\textwidth}

\textbf{Purpose:} Check if swapping the order of names affects the head's attention pattern

\begin{lstlisting}[language=Python]
mi.attention_map_generation(["Jennifer and Stephen went to the station. Stephen gave a necklace to", "Kenneth and Ryan went to the house. Ryan gave a ring to"], query_positions=[13, 13], layer_head_pairs=[(9, 9)])
\end{lstlisting}

\textbf{Purpose:} Test with more complex scenarios - three people mentioned

\begin{lstlisting}[language=Python]
mi.attention_map_generation(["Tom, Jerry and Mike went shopping. Mike gave a gift to", "Alice, Bob and Carol were talking. Carol gave a letter to"], query_positions=[13, 13], layer_head_pairs=[(9, 9)])
\end{lstlisting}

\textbf{Purpose:} Verify logit lens output for the new test prompts

\begin{lstlisting}[language=Python]
mi.logit_lens(["Sarah and Michael went to the park. Sarah talked to", "Jennifer and Stephen went to the station. Stephen gave a necklace to"], token_positions=[10, 13], layer_head_pairs=[(9, 9)], top_k=10)
\end{lstlisting}
\end{minipage}
\end{agentmsg}

\subsection{Final Hypothesis}
\label{ap:final_hypothesis}

The agent's final hypothesis consists of (1) a summary of the supporting evidence and (2) a brief free-text description of the component's functional role. 

\begin{hypothesisbox}
\textbf{Layer 9, Head 9 functions as a specialized "human recipient identification" mechanism that operates specifically in social interaction contexts.}

The component implements a syntactic-semantic rule that activates when:
\begin{enumerate}
    \item The current token is "to" following an action verb (give, hand, send, etc.)
    \item The context contains human names mentioned earlier in the prompt
    \item One of those humans is performing the current action
\end{enumerate}

When these conditions are met, the head:
\begin{itemize}
    \item \textbf{Attends strongly} (typically 0.7--0.95 attention weight) to the human name that is NOT the current actor/subject
    \item \textbf{Promotes that name} as the likely next token through its output (logit lens shows near 1.0 probability)
    \item \textbf{Prioritizes} the first-mentioned non-actor when multiple candidates exist
\end{itemize}

\textbf{Key mechanistic properties:}
\begin{itemize}
    \item The head contributes \textbf{positively} to correct predictions (ablation reduces probability by 0.08--0.16)
    \item It encodes \textbf{positional/structural information} rather than semantic identity (patching doesn't transfer specific names)
    \item It's \textbf{specialized for human names} and fails to activate for non-human entities (companies, celestial bodies, etc.)
    \item It implements a \textbf{discourse coherence pattern} for reciprocal human interactions (if X acts, Y receives)
\end{itemize}

This specialization makes evolutionary sense as human social interactions follow predictable reciprocal patterns that are important for language models to capture. The head essentially implements the heuristic: ``in human social situations, actions directed `to' someone typically target the other person mentioned who hasn't acted yet.''

\tcbline

\textbf{Summary:} Layer 9, Head 9 identifies and promotes human names as recipients of actions by attending to the non-actor from previously mentioned people when positioned at ``to'' following action verbs, implementing a specialized syntactic pattern for predicting human social interactions with high accuracy across diverse prompts.
\end{hypothesisbox}

\section{Example of Interpretive Ambiguity: The Entity Tracking Task}
\label{ap:in-comp}

In the Entity Tracking task, a large group of heads is labeled as \emph{value fetcher} heads and is described as primarily responsible for retrieving the value of the correct object to the final token position.

However, when examining disagreements between the systems’ descriptions and the expert explanations, we identified several heads that do not clearly follow the behavior described in the original paper. To investigate this further, we ran three types of analyses on all heads in this group across 500 examples:

\begin{enumerate}
    \item \textbf{Attention analysis.} We measured the fraction of cases in which the token receiving the highest attention from the head corresponds to the correct object. For heads in this group, we expect the highest attention to be directed primarily toward the correct object.
    \item \textbf{Logit lens analysis.} We applied a logit lens to the output of the attention head and measured how often the top-predicted token corresponds to an \emph{incorrect} object mentioned in the prompt. For heads in this group, we expect incorrect objects to receive low probability.
\item \textbf{Counterfactual patching analysis.} We replaced the head’s value with the corresponding value from a counterfactual prompt in the dataset and measured how often this intervention increased the probability of the correct object by at least 1\%. The intuition behind this test is that if patching the head increases the probability of the correct answer, then the head’s original contribution was suppressing that answer; in other words, altering or ablating the head improves the model’s prediction of the correct object. For heads in this group, we therefore expect that ablating them would \emph{decrease} the probability of the correct object.
\end{enumerate}

We present the results in Table~\ref{tab:value_fetcher_heads}. For each type of experiment, we highlight the 10 heads whose behavior is least aligned with the expected behavior of heads in this group. As can be seen, some of the heads exhibit behavior that is not aligned with the expected behavior of a value-fetcher head.

\begin{table}[ht]
\centering

\begin{tabular}{lccc}
\toprule
\textbf{(Layer, Head)} & \textbf{Correct Object Attention ($\uparrow$)} & \textbf{Incorrect Object Prediction ($\downarrow$)} & \textbf{Counterfactual Patching} ($\downarrow$) \\
\midrule
(15, 13) & 0.73 & 0.04 & 0.05 \\
(21, 3)  & 0.72 & 0.26 & 0.03 \\
(24, 5)  & \textbf{0.17} & \textbf{0.66} & \textbf{0.18} \\
(20, 14) & 0.32 & 0.44 & 0.11 \\
(18, 8)  & 0.72 & 0.23 & 0.09 \\
(29, 7)  & 0.28 & \textbf{0.61} & 0.09 \\
(18, 3)  & 0.89 & 0.02 & 0.04 \\
(15, 18) & 0.65 & 0.00 & 0.04 \\
(17, 28) & 0.78 & 0.04 & 0.06 \\
(21, 4)  & 0.28 & \textbf{0.48} & \textbf{0.15 }\\
(21, 25) & 0.20 & \textbf{0.62} & \textbf{0.15} \\
(23, 15) & 0.36 & \textbf{0.59} & 0.07 \\
(18, 28) & 0.53 & 0.01 & 0.05 \\
(23, 19) & \textbf{0.23} & \textbf{0.48} & \textbf{0.15} \\
(23, 20) & 0.24 & \textbf{0.59} & \textbf{0.17} \\
(19, 30) & 0.72 & 0.16 & 0.10 \\
(23, 5)  & \textbf{0.21} & 0.33 & \textbf{0.17} \\
(17, 27) & 0.32 & 0.12 & 0.04 \\
(15, 5)  & 0.90 & 0.00 & \textbf{0.15} \\
(21, 0)  & \textbf{0.22} & \textbf{0.47} & \textbf{0.17} \\
(23, 17) & \textbf{0.15} & 0.23 & 0.12 \\
(15, 2)  & 0.65 & 0.04 & 0.10 \\
(17, 3)  & 0.49 & 0.00 & 0.06 \\
(19, 20) & 0.38 & 0.13 & 0.13 \\
(19, 11) & 0.35 & 0.29 & 0.10 \\
(19, 8)  & 0.36 & 0.34 & 0.14 \\
(15, 6)  & 0.75 & 0.11 & 0.12 \\
(20, 29) & \textbf{0.17} & 0.10 & 0.10 \\
(16, 23) & 0.31 & 0.06 & 0.07 \\
(24, 0)  & \textbf{0.13} & 0.28 & \textbf{0.16} \\
(25, 14) & \textbf{0.20} & 0.17 & 0.14 \\
(14, 13) & \textbf{0.02} & 0.00 & 0.11 \\
(21, 26) & 0.27 & \textbf{0.46} & \textbf{0.15} \\
(24, 8)  & 0.25 & 0.20 & 0.14 \\
(18, 6)  & 0.51 & 0.19 & 0.09 \\
(19, 26) & 0.42 & 0.23 & 0.09 \\
(23, 16) & \textbf{0.11} & 0.35 & 0.13 \\
(16, 27) & 0.49 & 0.13 & 0.09 \\
(18, 20) & 0.43 & 0.27 & 0.09 \\
(18, 25) & 0.28 & 0.13 & 0.10 \\
\bottomrule
\end{tabular}
\caption{Evaluation of heads labeled as \emph{value fetcher} heads in the Entity Tracking task across three analyses on 500 examples. \textbf{Correct Object Attention} measures the fraction of cases in which the head attends most strongly to the correct object token (higher is better). \textbf{Incorrect Object Prediction} measures the fraction of cases in which applying a logit lens to the head’s output assigns the highest probability to an incorrect object mentioned in the prompt (lower is better). \textbf{Counterfactual Patching Effect} measures the fraction of cases in which replacing the head’s value with that from a counterfactual prompt increases the probability of the correct object by at least 1\% (lower is better). For each analysis, we highlight the 10 heads whose behavior is least aligned with the expected behavior of value fetcher heads.}
\label{tab:value_fetcher_heads}
\end{table}

\section{Example Of Memorization}
\subsection{IOI}
\label{ap:ioi_memory}

Here we provide examples illustrating the extent to which Claude Opus~4.1 and GPT-5 appear to have memorized the IOI task. Given the circuit name and a single task example, Claude is able to recall all previously identified functional roles along with their associated attention heads. The response was generated via an API call and without access to the web:

\begin{pythoncode} [title=User Prompt, breakable]
    You are an AI assistant with expertise in mechanistic interpretability research. You will be given a prompt asking about technical details of neural network circuits.

<prompt>
You are a mechanistic interpretability researcher. Tell us everything you know about the IOI (Indirect Object Identification) circuit.

For example, in sentences like: "When Mary and John went to the store, John gave a drink to"

In your answer, please mention:
- What attention heads are part of the circuit (e.g., layer 3, head 2 or L3H2). Mention all the heads you know of.
- What role each attention head plays in the circuit.
- How the circuit works step-by-step to predict the correct indirect object
</prompt>

Before responding, use the scratchpad below to organize your knowledge and ensure you cover all the requested points systematically.

<scratchpad>
Think through what you know about the topic in the prompt. Organize your knowledge into the specific categories requested. Make sure you have concrete details like specific layer and head numbers, step-by-step explanations, and research findings before writing your final response.
</scratchpad>

Now provide a comprehensive response that addresses all aspects mentioned in the prompt. Your response should be well-organized and technically accurate, drawing from your knowledge of mechanistic interpretability research. Structure your answer to clearly address each point requested in the prompt.

If you don't have specific information about certain technical details mentioned in the prompt, acknowledge this rather than speculating.

Write your response inside <answer> tags.
\end{pythoncode}
\begin{pythoncode}[title=Claude Response, breakable]
<scratchpad>
The IOI (Indirect Object Identification) circuit is a well-studied circuit in GPT-2 small that was discovered and analyzed by Wang et al. (2022) in their paper "Interpretability in the Wild". Let me organize what I know:

1. The task: Predicting the indirect object in sentences with two names, where one name appears twice (the repeated name should be predicted as the indirect object)

2. Key attention heads involved:
- Name Mover heads: These copy the correct name to the final position
- S-Inhibition heads: These suppress the subject (wrong answer)
- Duplicate Token heads: Identify which token appears twice
- Previous Token heads: Attend from the final position to previous tokens
- Induction heads: Help with the copying pattern

3. The circuit mechanism involves multiple stages of processing with different heads playing specific roles

Let me recall the specific layer/head numbers from the research...
</scratchpad>

<answer>
The IOI (Indirect Object Identification) circuit is a sophisticated mechanism discovered in GPT-2 small that solves the task of predicting the correct indirect object in sentences with a specific structure. This circuit was comprehensively analyzed by Wang et al. (2022) in their landmark paper "Interpretability in the Wild."

## The Task
The IOI task involves sentences where two names appear, with one name repeated, and the model must predict which name comes after "to" at the end. For example: "When Mary and John went to the store, John gave a drink to [Mary]" - where Mary is the correct prediction (the indirect object).


## Key Attention Heads in the Circuit
The circuit consists of several types of specialized attention heads:
### Name Mover Heads
- **L9H9, L9H6, L10H0** (among others in layers 9-11)
- Role: These heads directly copy the correct name from its earlier position to the final token position, enabling the model to output the correct indirect object

### S-Inhibition Heads (S-Inhibition = Subject Inhibition)
- **L7H3, L7H9, L8H6, L8H10**
- Role: These heads suppress attention to the subject (the repeated name that is NOT the answer), preventing the model from incorrectly predicting the subject

### Duplicate Token Heads
- **L0H1, L0H10, L1H4, L2H2, L3H0**
- Role: These early-layer heads identify which name appears twice in the sequence, marking it as the subject to be avoided

### Previous Token Heads
- **L2H2, L4H11** (among others)
- Role: These heads attend from each position to the previous token, helping to route information through the circuit

### Induction Heads
- **L5H5, L5H8, L5H9, L6H9**
- Role: These implement a copying pattern, attending from the second occurrence of a name back to tokens following its first occurrence

\end{pythoncode}

\begin{pythoncode}[title=GPT-5 Response, breakable]
<answer>
Overview
The IOI (Indirect Object Identification) circuit is a compact, largely attention-driven subcircuit in GPT-2-small that lets the model predict the correct indirect object in sentences like "When Mary and John went to the store, John gave a drink to ...". The model needs to output the other name (Mary) rather than the subject (John). The circuit has three main pieces:
- Name mover heads late in the model that copy name information to the prediction site and directly write to name logits.
- S-inhibition heads in the middle that identify the subject name and actively suppress its logit at the final position.
- Generic copy/induction/previous-token heads earlier that propagate name identity features so the later heads have something to work with.

What heads are part of the circuit (indices and roles)
Head indices can vary slightly across runs and checkpoints; the ones below are the canonical heads reported for GPT-2-small in the Anthropic "A Circuit for Indirect Object Identification in GPT-2" work and subsequent replications (e.g., in TransformerLens tutorials). I list the ones I know with their roles:

Late "name mover" and related heads (logit writers)
- L9H6 and L9H9 --- Primary name mover heads.
  Role: From the final position (after "to"), they attend strongly to the two name tokens and copy their value vectors into the residual stream at the final token. Because of how their OV circuits align with the unembedding, these heads directly write positive logit to the IOI name and not to the subject name (or write less/negative to the subject after upstream editing).
- L10H10 --- Backup name mover head.
  Role: Redundant with the L9 name movers; kicks in when the primaries are disrupted and helps maintain IOI performance.
- L10H7 --- Negative name mover (subject suppressor).
  Role: Attends to the subject name and writes a value that is anti-aligned with the subject's unembedding direction at the final position, directly decreasing the subject name logit.

Middle "S-inhibition" heads (subject detectors/suppressors)
- L7H6 and L7H9 --- S-inhibition heads.
  Role: Identify which of the two names is the subject of the main verb and write a negative contribution (via OV) targeted at that name's logit at the final position. They also shape the context so that the name movers preferentially boost the IOI rather than the subject.

Early and middle generic copy/position/induction heads (supporting infrastructure)
These are not IOI-specific but are used by the IOI circuit to propagate name identity and local syntax:
- Induction heads (canonical examples in GPT-2-small): L5H8, L6H9.
  Role: Copy token features forward (n-gram continuation), preserving name identity and helping the model keep both names' representations accessible at later positions.
- Previous-token heads (very early): examples include heads in L0 like L0H1, L0H10 (indices depend on run).
  Role: Provide short-range copying/positional context so that names and local function words remain represented.
- Other generic heads (various layers) that detect punctuation/conjunctions ("and", commas) and clause boundaries; their exact identities vary and they are not uniquely IOI-specific, so I won't list exact indices I'm not confident about.

Step-by-step: how the circuit predicts the correct indirect object
1) Early layers (L0--L3): Build local token/position features and preserve name identity
- Previous-token and short-range heads copy/align adjacent context so that name tokens, conjunction "and", punctuation, and the main verb region are cleanly represented.
- The two names' identity features are encoded at their positions and kept alive in the residual stream.

2) Middle layers (L4--L7): Maintain and disambiguate the two names; compute subjectness; start suppressing the subject
- Induction heads (e.g., L5H8, L6H9) propagate name identities forward and keep the two name representations accessible from later positions.
- S-inhibition heads (L7H6, L7H9) use syntactic patterning (e.g., which name is near the main verb as subject, the "gave ... to" frame, commas/conjunction structure) to determine which name is the subject.
- These S-inhibition heads then write a negative component into the residual stream at or routed to the final position, specifically anti-aligning with the subject name's unembedding vector. This pre-emptively dampens any subject logit that later heads might add.

3) Late layers (L8--L10): Copy names to the final token and write logits
- The primary name mover heads (L9H6, L9H9) at the final token attend to the positions of both names. Through their OV circuits, they copy the name value vectors to the final position. Because S-inhibition has already marked the subject name for suppression and because of the name movers' learned value/unembedding alignment, the IOI name's contribution is strongly positive, while the subject name gets little or negative net contribution.
- The negative name mover (L10H7) further suppresses the subject name's logit if needed.
- The backup name mover (L10H10) adds redundancy so the correct IOI logit boost is present even if one primary pathway is weak or ablated.

4) Output
- By the end of L9/L10, the logit lens shows a sharp logit gap favoring the IOI name over the subject. MLPs play a relatively minor role compared to these heads (the circuit is predominantly attention-mediated).
- The softmax over the vocabulary thus selects the IOI name.
</answer>
\end{pythoncode}

\subsection{Entity Tracking}
\label{ap:et_memory}
We probed Claude Opus~4.1 multiple times via API calls with the circuit name and a single task example to assess whether it could explicitly recall the conclusions of \citet{prakash2024fine}. In all runs, the model stated that it did not recognize the task and could not report the paper’s findings. Nevertheless, it often produced a high-level characterization of the model’s behavior that was broadly consistent with the paper’s conclusions:

\begin{pythoncode}[title=User Prompt,breakable]
You are an AI assistant with expertise in mechanistic interpretability research. You will be given a prompt asking about technical details of neural network circuits.

<prompt>
You are a mechanistic interpretability researcher. Tell us everything you know about the Entity Tracking circuit.

For example, in sentences like: "The flower is in Box U, the disk is in Box K, the bread is in Box Q, the radio is in Box Z, the drink is in Box F, the leaf is in Box L, the block is in Box C. Box Z contains the"

In your answer, please mention:
- What attention heads are part of the circuit (e.g., layer 3, head 2 or L3H2). Mention all the heads you know of.
- What role each attention head plays in the circuit.
- How the circuit works step-by-step to predict the correct indirect object
</prompt>

Before responding, use the scratchpad below to organize your knowledge and ensure you cover all the requested points systematically.

<scratchpad>
Think through what you know about the topic in the prompt. Organize your knowledge into the specific categories requested. Make sure you have concrete details like specific layer and head numbers, step-by-step explanations, and research findings before writing your final response.
</scratchpad>

Now provide a comprehensive response that addresses all aspects mentioned in the prompt. Your response should be well-organized and technically accurate, drawing from your knowledge of mechanistic interpretability research. Structure your answer to clearly address each point requested in the prompt.

If you don't have specific information about certain technical details mentioned in the prompt, acknowledge this rather than speculating.

Write your response inside <answer> tags.

\end{pythoncode}

\begin{pythoncode}[title=Claude Response,breakable]
<scratchpad>
Let me think through what I know about entity tracking circuits in language models:

1. Entity tracking involves keeping track of which entities are associated with which properties or locations
2. In the example given, the model needs to track which item is in which box, then retrieve the correct item when asked about a specific box
3. This requires:
   - Identifying entity-location pairs
   - Storing these associations
   - Retrieving the correct entity when given a location

For attention heads, I need to be careful - I don't have specific confirmed information about exact head numbers for an "Entity Tracking circuit" as described. Different papers may have studied similar phenomena under different names or in different models.

I should acknowledge what I don't know specifically while explaining the general mechanics of how such a circuit would need to work.
</scratchpad>

<answer>
I should note upfront that I don't have specific, confirmed information about a circuit explicitly called the "Entity Tracking circuit" with definitively identified attention head numbers. However, I can explain how such a circuit would need to function based on mechanistic interpretability principles and related research on similar tasks.

## How the Circuit Would Work

For the task you've described (tracking which items are in which boxes), the circuit would need to perform several key operations:

### Step 1: Entity-Location Binding
Early attention heads would need to identify and bind entities (flower, disk, bread, etc.) with their corresponding locations (Box U, Box K, Box Q, etc.). These heads would likely attend from the box identifier tokens to the entity tokens within the same clause.

### Step 2: Information Storage
Middle layer heads would need to copy or move this binding information to positions where it can be accessed later. This might involve:
- Copying entity information to the box name tokens
- Creating persistent representations that maintain the entity-location associations

### Step 3: Query Processing
When the model encounters "Box Z contains the", attention heads would need to:
- Identify that "Box Z" is a query for a stored entity
- Look back through the context to find where "Box Z" was mentioned with its contents

### Step 4: Entity Retrieval
Late-layer attention heads would need to:
- Attend from the final token position to the location where "radio is in Box Z" appears
- Copy the entity information ("radio") to enable prediction

## Circuit Components

While I cannot provide specific layer/head numbers for this exact circuit, similar circuits for tracking and retrieval tasks typically involve:
- **Early heads (layers 0-3)**: Local syntactic processing and initial entity recognition
- **Middle heads (layers 4-7)**: Information movement and storage
- **Late heads (layers 8-11)**: Retrieval and final answer composition

## Related Research

This type of circuit shares similarities with:
- Indirect object identification circuits studied in GPT-2
- Copying circuits that move information across sequence positions
- Memory circuits that store and retrieve factual associations

\end{pythoncode}
\section{Noise Sensitivity Analysis}
\label{ap:noise}
To assess how strongly the systems rely on experimental evidence, we introduce a noise parameter $\alpha$ into the execution of the interpretability tools requested by the agent. This parameter controls the amount of noise injected into the experimental results.

For each tool, we generate a “noised” version of the original results by randomly permuting them. We then construct a noisy result using a noise level $\alpha \in [0,1]$:
\[
(1 - \alpha)\cdot \text{original\_result} + \alpha \cdot \text{permuted\_result}.
\]

For each tool, we define the \emph{original result} as follows:
\begin{itemize}
    \item \textbf{Logit Lens.}  
    The original result is the probability vector obtained by applying the logit lens to a given component’s output.

    \item \textbf{Patching.}  
   The original result is the vector of token-wise probability differences between the original and patched runs.

    \item \textbf{Attention Patterns.}  
    The original result consists of the attention weight distribution produced by a given attention head.
\end{itemize}

\begin{figure}[h]
    \centering
    \includegraphics[width=0.4\textwidth]{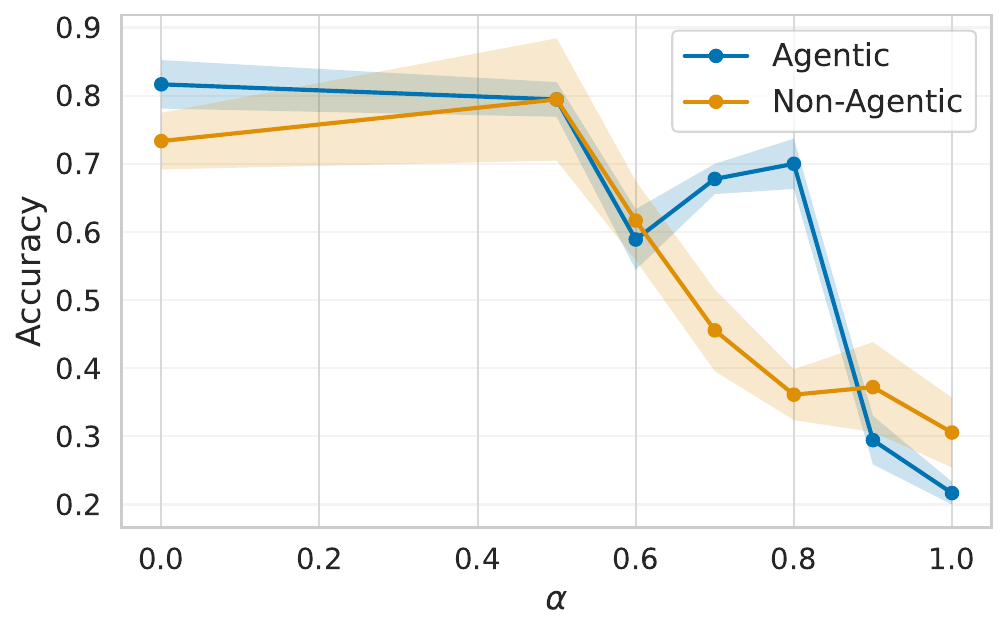}
    \includegraphics[width=0.4\textwidth]{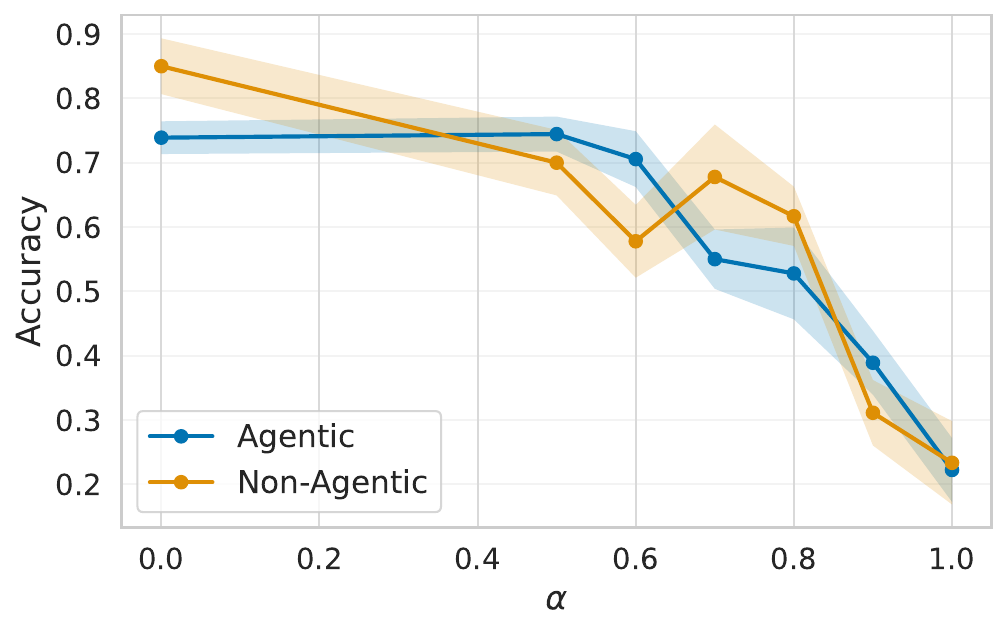}
    \caption{Each point in the graph represents the mean component functionality accuracy, averaged across all components in the IOI task, at a given noise level $\alpha$. We report results for $\alpha \in [{0, 0.5, 0.6, 0.7, 0.8, 0.9, 1}]$. Each plot corresponds to a different random seed, which determines both the specific task examples provided to the systems and the permutations applied to the experimental results. As shown, both systems exhibit clear performance degradation as the noise level increases.}
    \label{fig:noise_exp_ap}
\end{figure}

Figure~\ref{fig:noise_exp_ap} shows the results of the noising experiments on the IOI task. Each graph represents a run with a different seed, which affects both the task examples provided to the systems and the noise added to the activations.  
\vspace{1cm}

\end{document}